\documentclass[10pt,twocolumn,letterpaper]{article}

\usepackage[pagenumbers]{cvpr} % To force page numbers, e.g. for an arXiv version

\usepackage[accsupp]{axessibility}
\usepackage{graphicx}
\usepackage{amsmath}
\usepackage{amssymb}
\usepackage{booktabs}
\usepackage{xcolor}
\usepackage{pifont}

\DeclareMathOperator*{\argmin}{arg\,min}

\usepackage[pagebackref,breaklinks,colorlinks]{hyperref}

\usepackage[capitalize]{cleveref}
\crefname{section}{Sec.}{Secs.}
\Crefname{section}{Section}{Sections}
\Crefname{table}{Table}{Tables}
\crefname{table}{Tab.}{Tabs.}

\def\cvprPaperID{12085} % *** Enter the CVPR Paper ID here
\def\confName{CVPR}
\def\confYear{2023}

\begin{document}

\title{FJMP: Factorized Joint Multi-Agent Motion Prediction over Learned Directed Acyclic Interaction Graphs}

\author{Luke Rowe\thanks{Corresponding author.}, Martin Ethier, Eli-Henry Dykhne, Krzysztof Czarnecki\\ 
School of Computer Science, University of Waterloo\\
{\tt\small \{l6rowe, methier, ehdykhne, krzysztof.czarnecki\}@uwaterloo.ca}\\
{\href{https://rluke22.github.io/FJMP}{\texttt{https://rluke22.github.io/FJMP}}}
% \and
}
\maketitle

%%%%%%%%% ABSTRACT
\begin{abstract}
   Predicting the future motion of road agents is a critical task in an autonomous driving pipeline. In this work, we address the problem of generating a set of scene-level, or joint, future trajectory predictions in multi-agent driving scenarios. To this end, we propose FJMP, a Factorized Joint Motion Prediction framework for multi-agent interactive driving scenarios. FJMP models the future scene interaction dynamics as a sparse directed interaction graph, where edges denote explicit interactions between agents. We then prune the graph into a directed acyclic graph (DAG) and decompose the joint prediction task into a sequence of marginal and conditional predictions according to the partial ordering of the DAG, where joint future trajectories are decoded using a directed acyclic graph neural network (DAGNN). We conduct experiments on the INTERACTION and Argoverse 2 datasets and demonstrate that FJMP produces more accurate and scene-consistent joint trajectory predictions than non-factorized approaches, especially on the most interactive and kinematically interesting agents. FJMP ranks 1st on the multi-agent test leaderboard of the INTERACTION dataset.
\end{abstract}

%%%%%%%%% BODY TEXT
\section{Introduction}
\label{sec:intro}
Multi-agent motion prediction is an important task in a self-driving pipeline, and it involves forecasting the future positions of multiple agents in complex driving environments. Most existing works in multi-agent motion prediction predict a set of marginal trajectories for each agent \cite{liang2020lanegcn, gu2021densetnt, ye2021tpcn, liu2021mmtransformer, gao2020vectornet, sapp2022multipath++, jia2022hdgt}, and thus fail to explicitly account for agent interactions in the future. This results in trajectory predictions that are not consistent with each other. For example, the most likely marginal prediction for two interacting agents may collide with each other, when in reality a negotiation between agents to avoid collision is far more likely. As scene-consistent future predictions are critical for downstream planning, recent work has shifted toward generating a set of scene-level, or joint, future trajectory predictions \cite{sun2022m2i, gilles2022thomas, casas2020ilvm, ngiam2021scenetransformer, chen2022scept, cui2021lookout}, whereby each mode consists of a future trajectory prediction for each agent and the predicted trajectories are consistent with each other. 

\begin{figure}
    \centering
    \includegraphics[width=0.95\columnwidth]{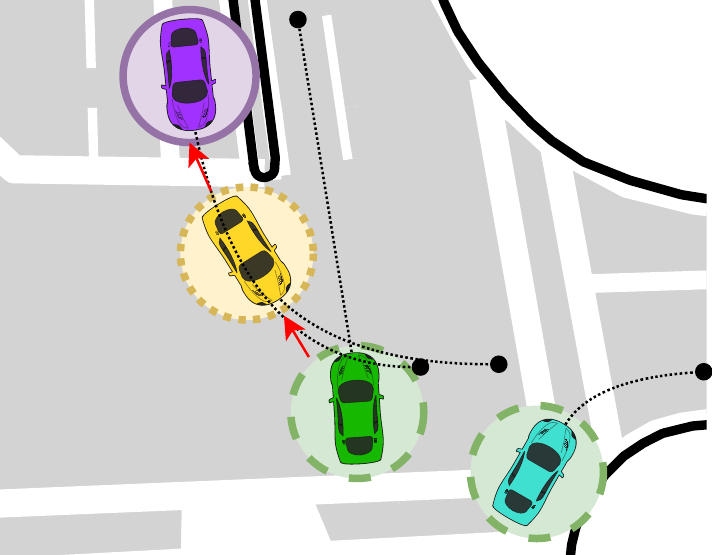}
    \caption{An illustration of the directed acyclic interaction graph, comprised of colored nodes and red arrows. The dotted black lines denote the ground-truth futures over a short time horizon. FJMP first produces marginal predictions for the green (dashed) nodes, followed by a conditional prediction for the yellow (dotted) node and a conditional prediction for the purple (solid) node, with conditioning on the predicted future of the parent nodes in the graph.}
    \label{fig:dag}
\end{figure}

In this work, we focus on the problem of generating a set of joint future trajectory predictions in multi-agent driving scenarios. Unlike marginal prediction, the joint trajectory prediction space grows exponentially with the number of agents in the scene, which makes this prediction setting particularly challenging. A common approach for this setting is to simultaneously predict the joint futures for all agents in the scene \cite{gilles2022thomas, casas2020ilvm, ngiam2021scenetransformer, girgis2022autobot, cui2021lookout}; however, this approach fails to explicitly reason about future interactions in the joint predictions. To address this limitation, recent work has shown that decomposing the joint prediction task of two interacting agents into a marginal prediction for the \textit{influencer} agent and a conditional prediction for the \textit{reactor} agent, where the reactor's prediction conditions on the predicted future of the influencer, can generate more accurate and scene-consistent joint predictions than methods that generate marginal predictions or simultaneous joint predictions \cite{sun2022m2i, li2022conditionalvectorized}. However, these methods are optimized for the joint prediction of only two interacting agents, and they do not efficiently scale to scenes with a large number of interacting agents. 

To address these limitations of existing joint motion predictors, in this work we propose \textbf{FJMP} -- a \textbf{F}actorized \textbf{J}oint \textbf{M}otion \textbf{P}rediction framework that efficiently generates joint predictions for driving scenarios with an arbitrarily large number of agents by \textit{factorizing} the joint prediction task into a sequence of marginal and conditional predictions. FJMP models the future scene interaction dynamics as a sparse directed interaction graph, where an edge denotes an explicit interaction between a pair of agents, and the direction of the edge is determined by their influencer-reactor relationship \cite{sun2022m2i, kumar2021hybrid, lee2019jointinteractiontrajectory}, as can be seen in \cref{fig:dag}. We propose a mechanism to efficiently prune the interaction graph into a directed acyclic graph (DAG). Joint future trajectory predictions are then decoded as a sequence of marginal and conditional predictions according to the partial ordering of the DAG, whereby marginal predictions are generated for the source node(s) in the DAG and conditional predictions are generated for non-source nodes that condition on the predicted future of their parents in the DAG. To enable this sequential trajectory decoding, we adapt a lightweight directed acyclic graph neural network (DAGNN) \cite{thost2021dagnn} architecture for efficiently processing predicted future information through the DAG and decoding the marginal and conditional trajectory predictions. Our main contributions can be summarized as follows:
\begin{itemize}
    \item We propose FJMP, a novel joint motion prediction framework that generates factorized joint trajectory predictions over sparse directed acyclic interaction graphs. To our knowledge, FJMP is the first framework that enables scalable factorized joint prediction on scenes with arbitrarily many interacting agents.
    \item We validate our proposed method on both the multi-agent INTERACTION dataset and the Argoverse 2 dataset and demonstrate that FJMP produces scene-consistent joint predictions for scenes with up to 50 agents that outperform non-factorized approaches, especially on the most interactive and kinematically complex agents. FJMP achieves state-of-the-art performance across several metrics on the challenging multi-agent prediction benchmark of the INTERACTION dataset and ranks 1st on the official leaderboard.
\end{itemize}

\section{Related Work}

\subsection{Motion Prediction in Driving Scenarios}

Given the recent growing interest in autonomous driving, many large-scale motion prediction driving datasets \cite{chang2019argoverse, wilson2021argoverse2, ettinger2021womd, caesar2020nuscenes, zhan2019interaction} have been publically released, which has enabled rapid progress in the development of data-driven motion prediction methods. Recurrent neural networks (RNNs) are a popular choice for encoding agent trajectories \cite{gilles2022thomas, gilles2021home, gilles2022gohome, deo2021lanegraphtraversals, kim2021lapred, chen2022scept} and convolutional neural networks (CNNs) are widely used in earlier works to process the birds-eye view (BEV) rasterized encoding of the High-Definition (HD) map \cite{girgis2022autobot, casas2020ilvm, park2020diverseandadmissible, narayanan2021divideandconquer, chen2022scept, gilles2021home}. As rasterized HD-map encodings do not explicitly capture the topological structure of the lanes and are constrained by a limited receptive field, recent methods have proposed vectorized \cite{gao2020vectornet, gu2021densetnt, liu2021mmtransformer, shi2022mtr}, lane graph \cite{liang2020lanegcn, gilles2022gohome, cui2021gorela}, and point cloud \cite{ye2021tpcn} representations for the HD-map encoding. Inspired by the success of transformers in both natural language processing \cite{vaswani2017transformer, devlin2019bert} and vision \cite{dosovitskiy2021vit}, several end-to-end transformer-based methods have recently been proposed for motion prediction \cite{girgis2022autobot, ngiam2021scenetransformer, liu2021mmtransformer, shi2022mtr, huang2022multimodaltransformer, zhou2022hivt, jia2022hdgt}. However, many of these transformer-based methods are extremely costly in model size and inference speed, which makes them impractical for use in real-world settings. FJMP adopts a LaneGCN-inspired architecture \cite{liang2020lanegcn} due to its strong performance on competitive benchmarks \cite{chang2019argoverse}, while retaining a small model size and fast inference speed.

\subsection{Interaction Modeling for Motion Prediction}

Data-driven methods typically use attention-based mechanisms \cite{liu2021mmtransformer, ngiam2021scenetransformer, girgis2022autobot, liang2020lanegcn, gilles2022thomas, zhou2022hivt} or graph neural networks (GNNs) \cite{casas2020ilvm, zeng2021lanercnn, li2021rain, liu2022isomorphism, gao2020vectornet, jia2022hdgt, kumar2021hybrid, casas2020spagnn} to model agent interactions for motion prediction. Recent works have demonstrated the importance of not only modeling agent interactions in the observed agent histories but also reasoning explicitly about the agent interactions that may occur in the future \cite{shi2022mtr, sun2022m2i, ban2022deepconceptnetwork, li2022conditionalvectorized, kuo2022linguistic, lee2019jointinteractiontrajectory}. MTR \cite{shi2022mtr} proposes to generate future trajectory hypotheses as an auxiliary task, where the future hypotheses are fed into the interaction module so that it can better reason about future interactions. Multiple works reason about future interactions through predicted pairwise influencer-reactor relationships \cite{sun2022m2i, li2022conditionalvectorized, lee2019jointinteractiontrajectory}, where the agent who reaches the conflict point first is defined as the influencer, and the reactor otherwise. FJMP uses attention to model interactions in the agent histories and constructs a sparse interaction graph based on pairwise influencer-reactor relationships to model future interactions. 

\begin{figure*}
  \centering
  \includegraphics[width=1\textwidth]{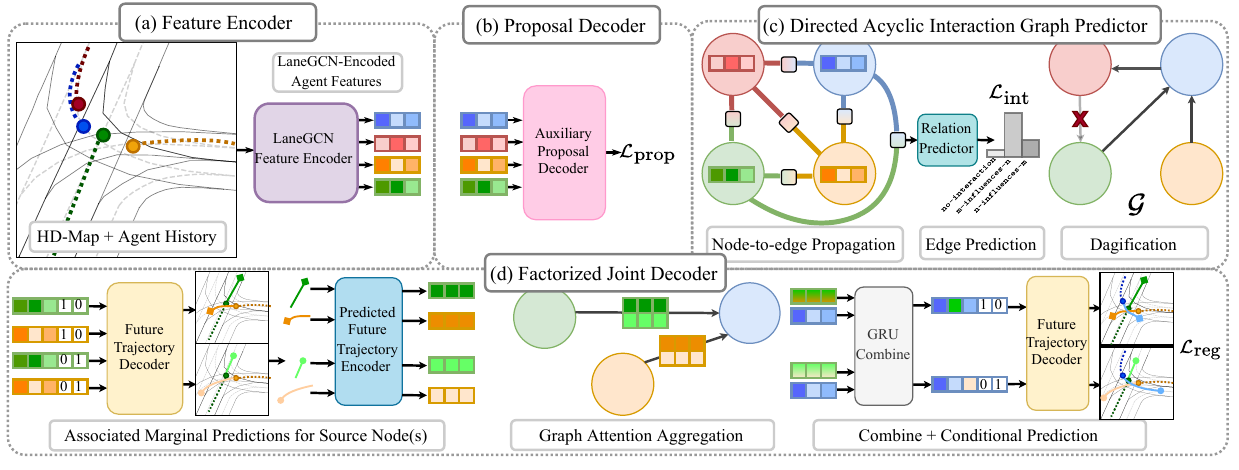}
    \caption{Illustration of the proposed FJMP framework. (a) Agent histories and the HD-Map are first processed by a LaneGCN-inspired feature encoder. (b) During training, the LaneGCN-encoded features are fed into an auxiliary future proposal decoder trained with a regression loss to encourage the LaneGCN features to be \textit{future-aware}. (c) The future-aware LaneGCN-features are processed by a GNN that predicts the pairwise influencer-reactor relationships supervised by a focal loss. A directed interaction graph $\mathcal{G}$ is constructed from the predicted edge probabilities and cycles are removed via an efficient ``dagification" procedure. (d). The predicted DAG and future-aware LaneGCN features are fed into a factorized DAGNN-based trajectory decoder (red agent removed for simplicity), which produces $K$ ($K = 2$ shown above) factorized joint futures in parallel and is supervised by a joint regression loss.}
    \label{fig:model}
\end{figure*}

\subsection{Joint Motion Prediction}

The majority of existing motion prediction systems generate marginal predictions for each agent \cite{sapp2022multipath++, gilles2022gohome, song2021prime, deo2021lanegraphtraversals, gao2020vectornet, park2020diverseandadmissible, narayanan2021divideandconquer, kim2021lapred, huang2022multimodaltransformer, gilles2021home, bhattacharyya2022ssllanes, ye2022dcms, liang2020lanegcn, gu2021densetnt, ye2021tpcn, zhou2022hivt, liu2021mmtransformer, wang2022ltp, jia2022hdgt, shi2022mtr, zeng2021lanercnn, cui2021gorela}; however, marginal predictions lack an association of futures across agents. Recent works have explored generating simultaneous joint predictions \cite{ngiam2021scenetransformer, casas2020ilvm, cui2021lookout, gilles2022thomas, girgis2022autobot}, but these methods do not explicitly reason about future interactions in the joint predictions. Other works generate joint predictions for two-agent interactive scenarios by selecting $K$ joint futures among all $K^2$ possible combinations of the marginal predictions \cite{shi2022mtr, wu2021air2}, which quickly becomes intractable as the number of agents in the scene increases. ScePT \cite{chen2022scept} proposes to handle the exponentially growing joint prediction space by decomposing joint prediction into the prediction of interactive cliques. However, the density of large cliques imposes a severe computational burden at inference time, which requires ScePT to upper-bound the maximum clique size to 4. To avoid the computational burden associated with dense interaction graphs, FJMP models future interactions as a \textit{sparse} interaction graph consisting only of the strongest interactions, which enables efficient joint decoding over interactive scenarios with many interacting agents.

Our proposed method is most closely related to M2I \cite{sun2022m2i}, which first predicts the influencer-reactor relationship between a pair of interacting agents and then generates a marginal prediction for the influencer agent followed by a conditional prediction for the reactor agent. However, we differ from M2I in three critical ways. First, M2I is designed specifically to perform joint prediction of two interacting agents, as their model design assumes one influencer agent and one reactor agent. In contrast, FJMP naturally scales to an arbitrary number of interacting agents, where an agent may have multiple influencers and influence multiple reactors. Second, M2I requires a costly inference-time procedure that does not scale to multiple agents whereby $N$ conditional predictions are generated for each marginal prediction, resulting in $N^2$ joint predictions that are pruned to $K = 6$ based on predicted likelihood. On the contrary, FJMP coherently aligns the joint predictions of a given modality through the DAG and directly produces $K = 6$ factorized joint predictions without any required pruning, which allows the system to seamlessly scale to scenes with an arbitrarily large number of agents. Third, M2I uses separate decoders for the marginal and conditional prediction, whereas FJMP decodes both marginal and conditional predictions using the same decoder, making it more parameter-efficient.

\section{FJMP}
\label{sec:fjmp}

In this section, we describe our proposed factorized joint motion prediction framework, illustrated in \cref{fig:model}.

\subsection{Preliminaries}
\label{subsec:preliminaries}

\subsubsection{Proposed Joint Factorization}
\label{subsubsec:proposedjoint}

The goal of multi-agent joint motion prediction is to predict the future $T_{\text{fut}}$ timesteps of $N$ dynamic agents in a scene given the past motion of the $N$ agents and the structure of the HD-Map. As there are multiple possible futures for a given past, the joint motion prediction task involves predicting $K > 1$ modalities, whereby each modality consists of a predicted future for each agent in the scene. We let $X$ and $Y$ denote the past trajectories and future trajectories for all $N$ agents in the scene, respectively, where $X_{\mathcal{S}}$ denotes the past trajectories for all agents in the set $\mathcal{S} \subseteq [N]$, and $Y_{\mathcal{S}}$ is defined similarly. Moreover, we let $C$ be an encoding of the HD-Map context.  

We first propose to model the future scene interaction dynamics as a DAG $\mathcal{G} = \{ \mathcal{V}, \mathcal{E} \}$, where the vertices $\mathcal{V} = [N]$ correspond to the $N$ dynamic agents in the scene, and a directed edge $e_{mn} \in \mathcal{E}$; $m,n \in [N]$, denotes an explicit interaction between agents $m$ and $n$ whereby $m$ is the influencer and $n$ is the reactor of the interaction. We propose to factorize the joint future trajectory distribution $P( Y | X, C)$ over the DAG $\mathcal{G}$ as follows:
\begin{align}
P(Y | X, C) = \prod_{n = 0}^{N-1} P(Y_{\{n\}} | Y_{\text{pa}_{\mathcal{G}}(n)}, X, C),
\end{align}
where $\text{pa}_{\mathcal{G}}(n)$ denotes the set containing the parents of node $n$ in $\mathcal{G}$. Intuitively, the proposed joint factorization can be interpreted as an inductive bias that encourages accounting for the predicted future of the agent(s) that influence agent $n$ when predicting the future of  agent $n$. We hypothesize that this inductive bias will ease the complexity of learning the joint distribution when compared to methods that produce a joint prediction for all $N$ agents simultaneously.

\subsubsection{Input Preprocessing}

The past trajectory of a given agent is expressed as a sequence of $T_\text{obs}$ states, which contains the 2D position, the velocity, and the heading of the agent at each timestep. We denote the past state of agent $n$ at timestep $t, t \in [T_\text{obs}]$, by $\mathbf{x}^n_t = [\mathbf{p}^n_t, \mathbf{v}^n_t, \psi^n_t]$, where $\mathbf{p}^n_t \in \mathbb{R}^2$ is the position, $\mathbf{v}^n_t \in \mathbb{R}^2$ is the velocity, and $\psi^n_t \in \mathbb{R}$ is the yaw angle. We are also provided the agent type $a^n$. As in LaneGCN~\cite{liang2020lanegcn}, we convert the sequence of 2D positional coordinates of each agent $n$ to a sequence of coordinate displacements: $\hat{\mathbf{p}}_t^n = \mathbf{p}^n_t - \mathbf{p}^n_{t-1}$ for all $t$. We encode the HD-Map as a lane graph with $M$ nodes, each denoting the location of the midpoint of a lane centerline segment. Using the lane graph construction proposed in LaneGCN, four adjacency matrices, $\{\mathbf{A}_i\}_{i \in \{ \text{pre, suc, left, right} \}}$, $\mathbf{A}_i \in \mathbb{R}^{M \times M}$, are calculated to represent the predecessor, successor, left, and right node connectivities in the lane graph, respectively. 
 Our system takes as input the $M$ lane node positional coordinates, the lane node connectivities $\{\mathbf{A}_i\}_{i \in \{ \text{pre, suc, left, right} \}}$, and the preprocessed agent history states $[\mathbf{\hat{p}}^n_t, \mathbf{v}^n_t, \psi^n_t]$ for all $n \in [N], t \in [T_{\text{obs}}]$. 

\subsection{Feature Encoder}
\label{subsec:featureencoder}

To encode the agent history and HD-map data, we employ a LaneGCN backbone \cite{liang2020lanegcn} with a few key modifications. For processing the agent histories, we replace LaneGCN's proposed ActorNet architecture with a gated recurrent unit (GRU) module. For processing the HD-Map, we employ the MapNet architecture, which consists of $L$ graph convolutional operators that enrich the lane node features by propagating them through the lane graph. We then employ the FusionNet architecture introduced in LaneGCN \cite{liang2020lanegcn} for fusing the map and actor features, but we remove the actor-to-lane (A2L) and lane-to-lane (L2L) modules, keeping only the lane-to-actor (L2A) and actor-to-actor (A2A) modules. We observed a minimal loss in performance when removing the A2L and L2L modules, and we benefited from the reduced parameter count. The output of the LaneGCN feature encoder produces a set of map-aware agent features $ H = \{ \mathbf{h}_n \}_{n \in [N]}$ for each agent.

\subsubsection{Auxiliary Proposal Decoder}
\label{subsubsec:auxiliaryproposaldecoder}

While the output of the LaneGCN feature encoder provides informative map-aware agent features, the A2A module only considers agent interactions in the observed past trajectories. However, these features will be used downstream to reason about agent interactions in the future, and thus we desire agent feature representations that are \textit{future-aware} -- agent features that are predictive of the future. To this end, we propose to regularize the LaneGCN agent feature representations with an auxiliary pretext task that predicts joint future trajectories on top of the LaneGCN-encoded agent features. We adopt a proposal decoder $f_{\text{prop}}$, which decodes $K$ joint future trajectories from the LaneGCN-encoded features $\{ \hat{\mathbf{y}}_{\text{prop}, k}^{n} \}_{k \in [K]} = f_{\text{prop}}(\mathbf{h}_n)$ and it is supervised by a joint regression loss $\mathcal{L}_{\text{prop}}$. More details of the joint regression loss can be found in \cref{app:lossfunctions}. We hypothesize that the proposed pretext task will regularize the LaneGCN feature representations, so that it contains future context that will be useful for reasoning about future interactions in the downstream modules. We note that the proposal decoder is discarded at inference time and is only used to regularize features during training.

\subsection{Directed Acyclic Interaction Graph Predictor}
\label{subsec:dagpredictor}

\subsubsection{Interaction Graph Predictor}
\label{subsubsec:interactiongraphpredictor}

In order to construct the directed acyclic interaction graph, we first must classify the future interaction label between every pair of agents in the scene. This task can be formulated as a classification task where we classify every edge in a fully-connected undirected interaction graph $\mathcal{G}_U = \{\mathcal{V}, \mathcal{E}_U \}$, where each agent corresponds to a node in $\mathcal{V}$.  Similar to \cite{sun2022m2i, lee2019jointinteractiontrajectory, kumar2021hybrid, li2022conditionalvectorized}, given an edge $e_{m,n} \in \mathcal{E}_U$, the classification task assumes three labels: \texttt{no-interaction}, \texttt{m-influences-n}, and \texttt{n-influences-m}, where the ground-truth future interaction label is heuristically determined using their ground-truth future trajectories. Concretely, we employ a collision checker to check for a collision between agents $m$ and $n$ at all pairs of future timesteps $(t_m, t_n)$ where $|t_m - t_n| \leq \epsilon_I$ for some threshold $\epsilon_I$. Details of the collision checker can be found in \cref{app:collisionchecker}. We let $\mathcal{C}$ denote the set of timestep pairs where a collision is detected. If $|\mathcal{C}| = 0$, then $m$ and $n$ are not interacting and the edge is labeled \texttt{no-interaction}. Otherwise, we identify the first such pair of timesteps $(\hat{t}_m, \hat{t}_n) \in \mathcal{C}$ where a collision is detected:
\begin{align}
    (\hat{t}_m, \hat{t}_n) &= \argmin_{(t_m, t_n) \in \mathcal{C}} \quad \min \{ t_m, t_n \}.
\end{align}
If $|\mathcal{C}| > 0$, then there exists a conflict point between the two agents, and the influencer agent is defined as the agent who reaches the conflict point first. Specifically, if $\hat{t}_m < \hat{t}_n$, then we assign the edge the label \texttt{m-influences-n}, and otherwise we assign the edge the label \texttt{n-influences-m}. 

With the heuristic interaction labels, we train a classifier to predict the interaction type on each edge of $\mathcal{G}_U$. We first initialize the node features of $\mathcal{G}_U$ to the future-aware LaneGCN agent features $\mathbf{h}_n$. We then perform a node-to-edge feature propagation step, where for each edge $e_{m,n}$:
\begin{align}
    \mathbf{h}^e_{m,n} = f_{\text{edge}}\Big( \left[ \mathbf{h}_m || \mathbf{h}_n || f_{\text{dist}}(\mathbf{p}^m_{t_\text{c}} - \mathbf{p}^n_{t_\text{c}}) || \mathbf{a}_{m,n} \right]\Big),
\end{align}
where $f_{\text{edge}}$ and $f_{\text{dist}}$ are 2-layer MLPs, $||$ denotes concatenation along the feature dimension, $t_\text{c}:=T_\text{obs} - 1$ is the present timestep, and $\mathbf{a}_{m,n} = f_{\text{type}}([a_m, a_n])$ is the output of a 2-layer MLP $f_{\text{type}}$ applied to the agent types $a_m, a_n$. We then classify the interaction label using a 2-layer MLP $f_{\text{int}}$ with a softmax activation:
\begin{align}
    \hat{r}_{m,n} = \text{softmax}(f_{\text{int}}(\mathbf{h}^e_{m,n})).
\end{align}
The interaction classifier is trained with a focal loss $\mathcal{L}_{\text{int}} = \mathcal{L}_{\text{focal}}^{\gamma, \alpha}(R, \hat{R})$ with hyperparameters $\gamma \text{ and } \alpha$, where $\hat{R}$ is the predicted interaction label distributions and $R$ is the ground-truth interaction labels. From the predicted interaction label distributions, we can construct a directed interaction graph $\mathcal{G} = \{\mathcal{V}, \mathcal{E}\}$ by selecting the interaction label on each edge with the highest predicted probability. For each pair of agents, we add a directed edge from the predicted influencer to the predicted reactor if an interaction is predicted to exist, and no edge is added otherwise.

% It is important to note that our proposed interaction labeling heuristic produces much sparser interaction graphs than the heuristic proposed in M2I \cite{sun2022m2i}. For each pair of agents, M2I's labelling heuristic BLAH if any pair of future trajectory coordinates in the agents' future trajectory horizon are within a threshold Euclidean distance of each other, where the threshold is taken to be the sum of the lengths of the two agents. However, for extremely congested scenes, this labeling heuristic , which makes it computationally expensive to perform factorized joint prediction over the graph. By restricting interactions to those that satisfy an explicit collision check, this ensures that interaction graph consists of only the strongest interactions. 

\begin{table*}[h]
\centering
\resizebox{0.65\textwidth}{!}{
\begin{tabular}{lc|ccccc}
\toprule
Model    & Venue     & minADE   & minFDE   & SMR   & CrossCol & CMR   \\ \midrule
THOMAS \cite{gilles2022thomas}   & ICLR 2022 & 0.416 & 0.968 & \textbf{0.179} & 0.128    & 0.252 \\
HDGT \cite{jia2022hdgt}     & -         & \underline{0.303} & \underline{0.958} & 0.194 & 0.163    & 0.236 \\
DenseTNT \cite{gu2021densetnt} & ICCV 2021 & 0.420  & 1.130  & 0.224 & \textbf{0.000}        & 0.224 \\
AutoBot \cite{girgis2022autobot}  & ICLR 2022 & 0.312 & 1.015 & 0.193 & 0.043    & 0.207 \\
HGT-Joint & - & 0.307 & 1.056 & 0.186 & 0.016 & 0.190 \\ 
Traj-MAE & - & 0.307 & 0.966 & \underline{0.183} & 0.021 & \underline{0.188} \\
\midrule
FJMP (Ours) & - & \textbf{0.275} & \textbf{0.922} & 0.185 & \underline{0.005} & \textbf{0.187} \\
\bottomrule 
\end{tabular}
}
\caption{Joint prediction results on the INTERACTION multi-agent \textbf{test} set. Methods are sorted by the official ranking metric (CMR). For each metric, the best method is \textbf{bolded} and the second-best method is \underline{underlined}. Lower is better for all metrics.}
\label{tab:interaction-test}
\end{table*}

%\subsubsection{Dagification}
%\label{subsubsec:dagification}

\textbf{Dagification.} In order to perform factorized joint prediction over the learned directed interaction graph $\mathcal{G}$, we require $\mathcal{G}$ to be a DAG. We propose to remove cycles from $\mathcal{G}$, or ``dagify" $\mathcal{G}$, by iterating through the cycles in $\mathcal{G}$ and removing the edges with the lowest predicted probability. We efficiently enumerate the cycles in $\mathcal{G}$ using Johnson's algorithm \cite{johnson1975cycle}, which has time complexity $O((|\mathcal{V}| + |\mathcal{E}|)(c + 1))$, where $c$ is the number of cycles in $\mathcal{G}$. As the directed interaction graphs are typically sparse $(|\mathcal{V}| \approx |\mathcal{E}|)$ with a small number of cycles, for our application Johnson's algorithm runs approximately linear in the number of agents in the scene.

\subsection{Factorized Joint Trajectory Decoder}
\label{subsec:factorizedjointtrajectorydecoder}

Given the future-aware LaneGCN feature encodings $ H = \{ \mathbf{h}_n \}_{n \in [N]}$ and the directed acyclic interaction graph $\mathcal{G}$, we perform factorized joint prediction according to the unique partial ordering of $\mathcal{G}$. We parameterize the factorized joint trajectory decoder using an adapted directed acyclic graph neural network (DAGNN) \cite{thost2021dagnn}. A DAGNN is a recently proposed architecture that is suited specifically for DAG classification tasks. The originally proposed DAGNN framework performs DAG-level classification tasks on top of the representations of the \textit{leaf} nodes in the DAG, where node features are propagated to the leaf nodes sequentially according to the partial ordering of the DAG. Although we desire to process the agents according to the partial ordering of the interaction graph $\mathcal{G}$, we also aim to use the intermediate updated node features of the DAG to generate conditional future trajectory predictions, and thus we adapt the DAGNN design to fit this criterion. We explain first how to produce a factorized joint prediction using the proposed adapted DAGNN decoder, and then how the proposed decoder is extended to produce multiple joint futures.

The factorized decoder first processes the source node(s) $\mathcal{S}$ in parallel. For each source node $s \in \mathcal{S}$, we first decode a marginal future trajectory prediction:
\begin{align}
    \mathbf{\hat{y}}^s &= \text{DECODE}(\mathbf{h}_s),
\end{align}
where $\text{DECODE}$ is a residual block followed by a linear layer and $\mathbf{\hat{y}}^s \in \mathbb{R}^{2 T_{\text{fut}}}$ is the sequence of predicted future trajectory coordinates. We then encode the predicted future trajectories of each source node $s \in \mathcal{S}$:
\begin{align}
    \mathbf{e}_s &= \text{ENCODE}(\mathbf{\hat{y}}^s),
\end{align}
where ENCODE is a 3-layer MLP. For each $s \in \mathcal{S}$, the encoding of the predicted future $\mathbf{e}_s$ is then fed along the outgoing edges of $s$. Namely, after processing the source nodes $\mathcal{S}$, we update the features of the nodes that are next in the partial ordering of $\mathcal{G}$. For every such node $n$, we perform the following update:
\begin{align}
\mathbf{h}_n &\leftarrow \text{COMB}\left( \text{AGG}\left(\{ \mathbf{e}_m + \mathbf{a}_{mn} | m \in \text{pa}_{\mathcal{G}}(n) \}\right), \mathbf{h}_n\right),
\end{align}
where AGG is a neural network that \textit{aggregates} the node features from $n$'s parents and COMB is a neural network that \textit{combines} this aggregated information with $n$'s features to update the feature representation of $n$ with conditional context about the predicted future of $n$'s parents. $\mathbf{a}_{mn} = f^{\text{dec}}_{\text{type}}([a_m, a_n])$ is the output of a 2-layer MLP $f^{\text{dec}}_{\text{type}}$ applied to the agent types. From here, we let $\mathbf{b}_{mn} := \mathbf{e}_m + \mathbf{a}_{mn}$. Similar to DAGNN \cite{thost2021dagnn} which uses additive attention, we parameterize AGG using graph attention \cite{petar2017gat}:
\begin{align}
    \mathbf{m}_n := \text{AGG}\left(\{ \mathbf{b}_{mn} | m \in \text{pa}_{\mathcal{G}}(n) \}\right) = \sum_{m \in \text{pa}_{\mathcal{G}}(n)} \alpha_{mn} \mathbf{W}_1 \mathbf{b}_{mn},
\end{align}
\begin{align}
    \alpha_{mn} = \frac{\exp(\text{LeakyReLU}(\mathbf{a}^{\top}\left[ \mathbf{W}_1 \mathbf{b}_{mn} || 
    \mathbf{W}_2 \mathbf{h}_n\right]))}{\sum_{k \in \text{pa}_{\mathcal{G}}(n)} \exp(\text{LeakyReLU}(\mathbf{a}^{\top}\left[ \mathbf{W}_1 \mathbf{b}_{kn} || 
    \mathbf{W}_2 \mathbf{h}_n\right]))}.
\end{align}
COMB is parameterized by a GRU recurrent module:
\begin{align}
    \mathbf{h}_n \leftarrow \text{COMB}(\mathbf{m}_n, \mathbf{h}_n) = \text{GRU}(\mathbf{m}_n, \mathbf{h}_n).
\end{align}
As the aggregated message $\mathbf{m}_n$ provides conditional context for updating the representation of $\mathbf{h}_n$, $\mathbf{m}_n$ is treated as the input and $\mathbf{h}_n$ is treated as the hidden state. It is important to note that the roles of the input and hidden state are reversed in the original DAGNN design \cite{thost2021dagnn}. The updated representation $\mathbf{h}_n$ for node $n$ is now imbued with conditional context about the predicted future of the parent(s) of $n$, which can now be fed into DECODE to produce conditional future predictions for agent $n$. We sequentially continue the process of encoding, aggregating, combining, and decoding according to the DAG's partial order until all nodes in the DAG have a future trajectory prediction. The future trajectory predictions of all nodes are then conglomerated to attain a factorized joint prediction. 

%\subsubsection{Multiple Futures}
%\label{subsec:multiplefutures}
\textbf{Multiple Futures.} To extend the DAGNN factorized decoder to produce multiple factorized joint predictions, we simply process $K$ copies of $ H = \{ \mathbf{h}_n \}_{n \in [N]}$ through the DAG in parallel. To ensure each copy of $H$ generates a different set of futures, we concatenate a one-hot encoding of the modality with $\mathbf{h}_n$ along the feature dimension, for each $n \in [N]$,  prior to it being fed into DECODE. This approach is similar to the multiple futures approach proposed in SceneTransformer \cite{ngiam2021scenetransformer} and we found it to work well in our application. We train the factorized joint predictor to produce diverse multiple futures by training with a winner-takes-all joint regression loss $\mathcal{L}_{\text{reg}}$. More details about the regression loss can be found in \cref{app:lossfunctions}.

\subsection{Training Details}
\label{subsection:trainingdetails}
We first train the interaction graph predictor separately using its own feature encoder weights. The interaction graph predictor is trained via gradient descent, where the loss function is defined by:
\begin{align}
    \mathcal{L}_1 = \mathcal{L}_{\text{int}} + \mathcal{L}_{\text{prop}}.
\end{align}
Next, we train the factorized joint decoder using its own feature encoder weights, where the interaction graphs $\mathcal{G}$ are generated with the trained interaction graph predictor. The factorized joint predictor is trained via gradient-descent, where the loss function is defined by:
\begin{align}
    \mathcal{L}_2 = \mathcal{L}_{\text{reg}} + \mathcal{L}_{\text{prop}}.
\end{align}
Similar to M2I \cite{sun2022m2i}, we employ teacher forcing of the influencer future trajectories during training, which helps to learn the proper influencer-reactor dynamics.

\begin{table*}[t]
\centering
\resizebox{\textwidth}{!}{
\begin{tabular}{lcc|cccc|cc|cc|cc}
\toprule
Dataset & Actors Evaluated & Model  & minFDE   & minADE   & SCR & SMR & iminFDE & iminADE & $\text{iminFDE}_3$ & $\text{iminADE}_3$ & $\text{iminFDE}_5$ & $\text{iminADE}_5$ \\ \midrule
 Interaction & - & Non-Factorized & 0.643 & 0.199 & 0.004 & 0.088 & 0.688   & 0.210   & 0.784    & 0.240   & 0.854   & 0.261   \\
 &  & FJMP  & 0.630 & 0.194 & 0.003 & 0.084 & 0.672  & 0.206   & 0.758   & 0.232   & 0.826   & 0.252 \\
 \cmidrule{3-13}
 &  & $\Delta$ & 0.013 & 0.005 & 0.001 & 0.004 & 0.016 & 0.004 & 0.026 & 0.008 & 0.028 & 0.009 \\
\midrule\midrule
Argoverse 2 & Scored & Non-Factorized & 1.965 & 0.834 & - & 0.349 & 2.957 & 1.223 & 3.276 & 1.340 & 3.436 & 1.399 \\
 &  & FJMP & 1.921 & 0.819 & - & 0.343 & 2.893 & 1.204 & 3.205 & 1.320 & 3.356 & 1.377\\
 \cmidrule{3-13}
 &  & $\Delta$ & 0.044 & 0.015 & - & 0.006 & 0.064 & 0.019 & 0.071 & 0.020 & 0.080 & 0.022\\
 \cmidrule{2-13}
& All & Non-Factorized  & 1.995 & 0.825 & - & 0.340 & 3.302 & 1.309 & 3.759 & 1.477 & 3.952 & 1.545\\
 &  & FJMP & 1.963 & 0.812 & - & 0.337 & 3.204 & 1.273 & 3.652 & 1.439 & 3.839 & 1.504\\
 \cmidrule{3-13}
  &  & $\Delta$ & 0.032 & 0.013 & - & 0.003 & 0.098 & 0.036 & 0.107 & 0.038 & 0.113 & 0.041\\
\bottomrule 
\end{tabular}
}
\caption{Non-Factorized Baseline vs. FJMP performance on joint metrics on the INTERACTION and Argoverse 2 validation sets. Lower is better for all metrics.
%We evaluate two settings on Argoverse 2: \textit{Scored}, which include both scored and focal agents in Argoverse 2 dataset; and \textit{All}, which includes all scored, focal, and unscored agents.
Argoverse 2 lacks agent bounding box information, so SCR is not computed. $\Delta$ denotes the difference in performance between FJMP and the Non-Factorized baseline.}
\label{tab:validation-results}
\end{table*}

\section{Experiments}
\label{sec:experiments}

%\subsection{Datasets}
%\label{subsec:datasets}
\textbf{Datasets.} We evaluate FJMP on the INTERACTION v1.2 multi-agent dataset and the Argoverse 2 dataset, as both have multi-agent evaluation schemes for scenes with many interacting agents and require predicting joint futures for scenes with up to 40 and 56 agents, respectively. However, currently, only INTERACTION has a public benchmark for multi-agent joint prediction. Argoverse 2 contains \textit{scored} and \textit{focal} actors, which are high-quality tracks near the ego vehicle; and \textit{unscored} actors, which are high-quality tracks more than 30\,m from the ego vehicle. We evaluate FJMP on (i) only the scored and focal actors; and (ii) all scored, focal, and unscored actors. More details about these datasets can be found in \cref{app:datasets}.

%\subsection{Evaluation Metrics}
%\label{subsec:evaluationmetrics}

\textbf{Evaluation Metrics.} We report the following joint prediction metrics: \textbf{minFDE} is the final displacement error (FDE) between the ground-truth and closest predicted future trajectory endpoint from the $K$ joint predictions; \textbf{minADE} is the average displacement error (ADE) between the ground-truth and closest predicted future trajectory from the $K$ joint predictions; \textbf{SMR} is the minimum proportion of agents whose predicted trajectories ``miss" the ground-truth from the $K$ joint predictions, where a miss is defined in \cref{app:missrate}; and \textbf{SCR} is the proportion of modalities where two or more agents collide. The INTERACTION test set additionally reports two joint prediction metrics: \textbf{CrossCol} is the same as SCR but does not count ego collisions, and \textbf{CMR} is the same as SMR but only considers modalities without non-ego collisions. For all metrics, we evaluate $K=6$. These six joint prediction metrics do not necessarily capture the performance on the most interactive and challenging cases in the dataset, which is critically important for benchmarking and improving motion prediction systems. To address this limitation, we propose two new interactive metrics: (i) \textbf{iminFDE} first identifies the modality $k$ with minimum FDE over all the agents in the scene and then computes the FDE of modality $k$ only over agents that are \textit{interactive}, which we heuristically define as agents with at least one incident edge in the ground-truth sparse interaction graph, where $\epsilon_{I} = 2.5$\,s. (ii) \textbf{iminADE} is defined similarly. We found that many of the interactive cases in the datasets contain kinematically simple cases where agents exhibit simple leader-follower behaviour. To evaluate the \textit{challenging} interactive cases, we further remove interactive agents in our evaluation that attain less than $d$ meters in FDE with a constant velocity model. These metrics are denoted $\textbf{iminFDE}_{\textbf{d}}$ and $\textbf{iminADE}_{\textbf{d}}$, where we report $d=3,5$. Please see \cref{app:constantvelocitymodel} for more details.

% INTERACTION, however, is the only driving dataset with a public benchmark for evaluating the joint prediction of more than 2 agents. 
%

%\subsection{Implementation Details}
%\label{subsec:implementationdetails}

\textbf{Implementation Details.} Our models are trained on 4 NVIDIA Tesla V100 GPUs using the Adam optimizer \cite{kingma2015adam}. The interaction graph predictor and factorized joint decoder are trained with the same hyperparameters. For INTERACTION, we set the batch size to 64 and train for 50 epochs with a learning rate of 1e-3, step-decayed by a factor of 1/5 at epochs 40 and 48. For Argoverse 2, we set the batch size to 128 and train for 36 epochs with a learning rate of 1e-3, step-decayed by a factor of 1/10 at epoch 32. As bounding-box information is not provided with Argoverse 2, the collision checker used to construct interaction labels uses a predefined length/width for each agent type, as listed in \cref{app:argoverse2dataset}. Our INTERACTION and Argoverse 2 models train in 10 and 15 hours. See \cref{app:trainingdetails} for more details.

%\subsection{Methods under Comparison}

\begin{table}[h]
\centering
\resizebox{1\columnwidth}{!}{
\begin{tabular}{l|ccccc}
\toprule
Model    & minFDE   & minADE  & SMR   & Prop. Edges & Inf. Time (s) \\ \midrule
Non-Factorized & 0.643 & 0.199 & 0.088 & - & \textbf{0.010} \\
FJMP (Dense)   & \textbf{0.623} & \textbf{0.193} & \textbf{0.081} & 0.180 & 0.062 \\
\midrule
FJMP    & \underline{0.626} & \textbf{0.193} & \underline{0.083} & 0.045 & \underline{0.038} \\

 \bottomrule 
\end{tabular}
}
\caption{Comparison of sparse vs. dense interaction graphs on the INTERACTION validation set. The FJMP model is trained and evaluated using the ground-truth sparse interaction graphs, and FJMP (Dense) is trained and evaluated using dense ground-truth interaction graphs attained via the M2I \cite{sun2022m2i} labeling heuristic. \textbf{Prop. Edges} measures the proportion of agent pairs connected in the ground-truth training interaction graphs. \textbf{Inf. Time} is the inference time per validation scene on 1 NVIDIA Tesla V100 GPU.}
\label{tab:sparse-dense}
\end{table}

\textbf{Methods under Comparison.} We compare FJMP against the top-performing methods on the INTERACTION multi-agent test set leaderboard \cite{gilles2022thomas, jia2022hdgt, gu2021densetnt, girgis2022autobot}. FJMP is the only method on the leaderboard that performs factorized joint prediction. To measure the improvement of FJMP over non-factorized approaches, we compare FJMP against a baseline called \textbf{Non-Factorized}, which computes $K$ simultaneous joint futures from the feature representations output by the proposed LaneGCN feature encoder. 

%\subsection{Results}

\textbf{Results.} The joint prediction results for the top-performing methods on the INTERACTION multi-agent test set are shown in \cref{tab:interaction-test}. FJMP performs the best on minFDE, minADE, and the official ranking metric CMR, while performing competitively on other metrics. Crucially, FJMP produces joint predictions that are both more accurate---as demonstrated by its superior performance on minADE and minFDE---and more scene-consistent---as demonstrated by its near-zero collision rate---than non-factorized approaches, which highlights the benefit of the proposed joint factorization.

\begin{table}[h]
\centering
\resizebox{\columnwidth}{!}{
\begin{tabular}{lcc|cc|cc}
\toprule
Model    & Prop? & TF? & minFDE   & minADE & iminFDE  & iminADE  \\ \midrule
Non-Factorized & \text{\ding{55}}  & \text{\ding{55}}  & 1.995 & 0.825 & 3.302 & 1.309 \\
FJMP     & \text{\ding{55}}  & \text{\ding{55}}  & 2.004 & 0.829 & 3.274 & 1.304 \\ 
FJMP     & \text{\ding{55}}  & \text{\ding{51}} & 2.001 & 0.827 & 3.300 & 1.312 \\
FJMP     & \text{\ding{51}} & \text{\ding{55}}  & \underline{1.987} & \underline{0.820} & \underline{3.255} & \underline{1.293} \\
FJMP     & \text{\ding{51}} & \text{\ding{51}} & \textbf{1.963} & \textbf{0.812} & \textbf{3.204} & \textbf{1.273}\\ \bottomrule
\end{tabular}
}
\caption{Ablation study of FJMP on Argoverse 2 validation set, All setting. \textbf{Prop?} denotes whether we include the proposal decoder during training. \textbf{TF?} denotes whether we teacher-force the influencer trajectories during training.}
\label{tab:argoverse2-ablation}
\end{table}

\Cref{tab:validation-results} reports validation results on the INTERACTION and Argoverse 2 datasets, where we compare FJMP against the baseline method without joint factorization. For Argoverse 2, we have two evaluation schemes: (i) we evaluate the joint predictions of the scored and focal agents (Scored), and (ii) we evaluate the joint predictions of the scored, unscored, and focal agents (All) to demonstrate its scalability to scenes with a large number of agents. The results show that the proposed joint factorized predictor consistently provides an improvement in performance over the non-factorized baseline. We expect that FJMP improves the most over the baseline on the interactive cases in the dataset, as the proposed factorization directly enables conditioning the reactor predictions on the predicted futures of their influencers. Importantly, we note that for scenes with no predicted interactions, the factorization becomes a product of marginal predictions and thus FJMP reduces to the non-factorized prediction. As expected, the relative improvement of FJMP over the baseline is larger on the interactive and kinematically interesting cases, as demonstrated by a larger performance improvement on the interactive minFDE/minADE metrics. This indicates that the performance improvement from the joint factorization concentrates on the challenging interactive cases, while still producing accurate joint predictions for the full scene. We refer readers to \cref{app:qualitativeresults} for qualitative examples demonstrating the benefit of factorized prediction.
% Figure BLAH illustrates a complex interactive scenario (DESCRIPTION TO BE PROVIDED HERE) where FJMP reasons explicitly about the future interactions to produce more accurate predictions than the non-factorized baseline.

\Cref{tab:sparse-dense} uses the INTERACTION dataset to ablate the design choice of representing the interaction graph sparsely with only the strongest pairwise interactions as edges in the graph. We compare FJMP against a variant of FJMP that uses a different labeling heuristic for the interaction graph, resulting in \textit{denser} interaction graphs. Namely, FJMP (Dense) uses the M2I \cite{sun2022m2i} heuristic: for each pair of agents, an interaction is defined to exist if any pair of future trajectory coordinates in the future trajectory horizon is within a threshold Euclidean distance of each other, where the threshold is taken to be the sum of the lengths of the two agents. The influencer-reactor relationship is determined by who reaches the conflict point first. We found that the M2I heuristic often adds several unnecessary edges, especially in congested scenes---as exemplified in \cref{app:sparsevsdense}. We train and evaluate the FJMP models in \cref{tab:sparse-dense} using the ground-truth interaction graphs to precisely compare the different labeling heuristics. The results show that the dense (M2I) interaction graph improves very slightly over the sparse interaction graph; however, we retain most of the improvement over the non-factorized baseline with the sparse interaction graph, which indicates that modeling only the strongest interactions is sufficient to see most of the improvement with joint factorization. Moreover, the sparse interactions contain  75\% fewer edges than the dense interaction graph, which accelerates inference by nearly 2x. 

\Cref{tab:argoverse2-ablation} conducts an ablation study on Argoverse 2 where we analyze the effect of using the auxiliary proposal decoder and teacher forcing of the influencer's future trajectories during training. The results indicate that both the auxiliary proposal decoder and teacher forcing is critical for allowing the model to reason appropriately about the influencer-reactor future dynamics. Notably, without the proposal decoder (rows 2 and 3 in \cref{tab:argoverse2-ablation}), FJMP performs similarly to the non-factorized baseline, which we hypothesize is because the LaneGCN-encoded features do not contain the necessary future context to reason appropriately about the future interactions. Teacher forcing also provides an additional performance benefit by removing the spurious noise in the predicted influencer trajectories, so that the model better learns the proper influencer-reactor dynamics during training. 

\section{Conclusion}
\label{sec:conclusions}

In this paper, we propose FJMP, a factorized joint motion prediction framework for multi-agent interactive driving scenarios. FJMP models the future scene interaction dynamics as a sparse directed acyclic interaction graph, which enables efficient factorized joint prediction. We demonstrate clear performance improvements with our factorized design on the Argoverse 2 and INTERACTION datasets and perform state-of-the-art on the challenging multi-agent INTERACTION benchmark. 

% \subsection{Limitations}
% \label{subsec:limitations}

\textbf{Limitations} The proposed framework adopts a heuristic labeling scheme to determine the ground-truth interaction graph. We observe a performance-efficiency tradeoff with a denser interaction graph; however, there may exist better heuristics for classifying future interactions that retain most of the sparsity of the interaction graph without trading off performance. Moreover, long chains of leader-follower behaviour in congested traffic may require costly sequential processing with our method. Finding mechanisms to prune the interaction graph to best trade-off performance and efficiency is a direction we plan to explore in future work.

\section{Acknowledgements}
\label{sec:acknowledgements}
This work was funded by Ontario Graduate Scholarship and NSERC. We thank Benjamin Thérien and Prarthana Bhattacharyya for their valuable insights and discussions.

% References
{\small
\bibliographystyle{ieee_fullname}
\bibliography{main}
}

\onecolumn
\appendix
\clearpage

\section{Loss Functions}
\label{app:lossfunctions}

The loss function for training the factorized joint trajectory decoder (\cref{subsec:factorizedjointtrajectorydecoder}) is defined by:
\begin{align}
    \mathcal{L}_2 = \mathcal{L}_{\text{reg}} + \mathcal{L}_{\text{prop}},
\end{align}
where $\mathcal{L}_{\text{reg}}(\{ \hat{Y}_{k} \}_{k \in [K]}, Y) := \mathcal{L}_{\ell_1}(\{ \hat{Y}_{k} \}_{k \in [K]}, Y)$ is a scene-level smooth $\ell_1$ regression loss applied to the best modality of $K=6$ joint modalities $\{ \hat{Y}_{k} \}_{k \in [K]}$, where the best modality attains the minimum loss:
\begin{align}
    \mathcal{L}_{\ell_1}(\{ \hat{Y}_{k} \}_{k \in [K]}, Y) = \min_{k \in [K]} \frac{1}{A \cdot T_{\text{fut}}} \sum_{a \in [A]} \sum_{t \in [T_{\text{fut}}]} \text{reg}(\hat{Y}_{t, k}^{a} - Y^{a}_t),
\end{align}
where $Y$ denotes the ground-truth future trajectory coordinates of all $A$ agents in the scene, $\text{reg}(\mathbf{x}) = \sum_i d(x_i)$, $x_i$ is the $i$'th element of $\mathbf{x}$, and $d(x)$ is the smooth $\ell_1$ loss defined by:
\begin{align}
    d(x) = \begin{cases}
        0.5x^2, & \text{if $||x||_1 \leq 1$}\\
        ||x||_1 - 0.5, & \text{otherwise.}
    \end{cases}
\end{align}
Similarly, the auxiliary decoder loss $\mathcal{L}_{\text{prop}}$ is a scene-level smooth $\ell_1$ loss applied to the best of $K=15$ joint proposals $\{ \hat{Y}^{\text{prop}}_{k} \}_{k \in [K]}$:
\begin{align}
    \mathcal{L}_{\text{prop}}(\{ \hat{Y}^{\text{prop}}_{k} \}_{k \in [K]}, Y) := \mathcal{L}_{\ell_1}(\{ \hat{Y}^{\text{prop}}_{k} \}_{k \in [K]}, Y).
\end{align}
We use the auxiliary proposal loss $\mathcal{L}_{\text{prop}}$ for training both the interaction graph predictor ($\mathcal{L}_1$ in \cref{subsection:trainingdetails}) and the factorized joint decoder ($\mathcal{L}_2$ in \cref{subsection:trainingdetails}) as both modules require explicit reasoning about interactions in the future trajectories, and thus \textit{future-aware} agent features are beneficial for both modules.

\section{FJMP System Diagram}
\label{app:systemdiagram}

\subsection{Training Time}
\label{app:trainingtime}

\begin{figure*}[h]
  \centering
  \begin{subfigure}{0.60\textwidth}
    \includegraphics[width=1\textwidth]{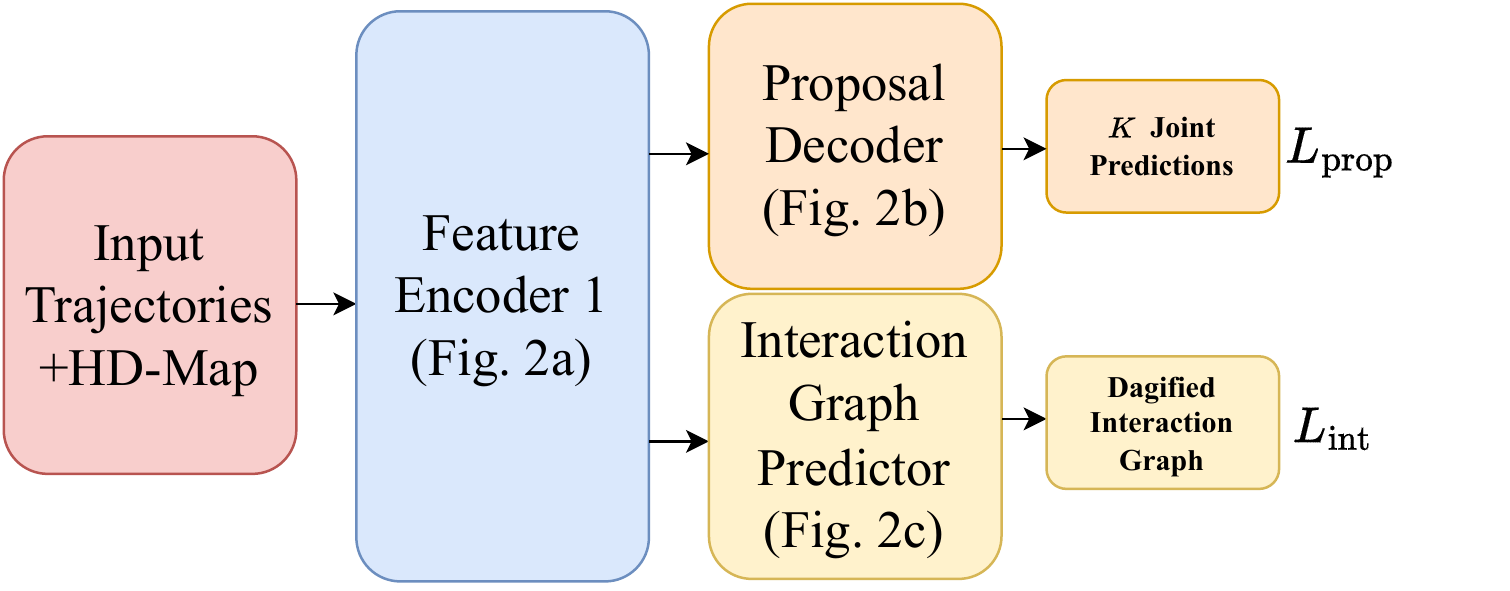}
    \caption{Interaction Graph Predictor training stage of FJMP.}
    \label{fig:trainingsystemdiagramstage1}
  \end{subfigure}
  \hfill
  \begin{subfigure}{0.39\textwidth}
    \includegraphics[width=1\textwidth]{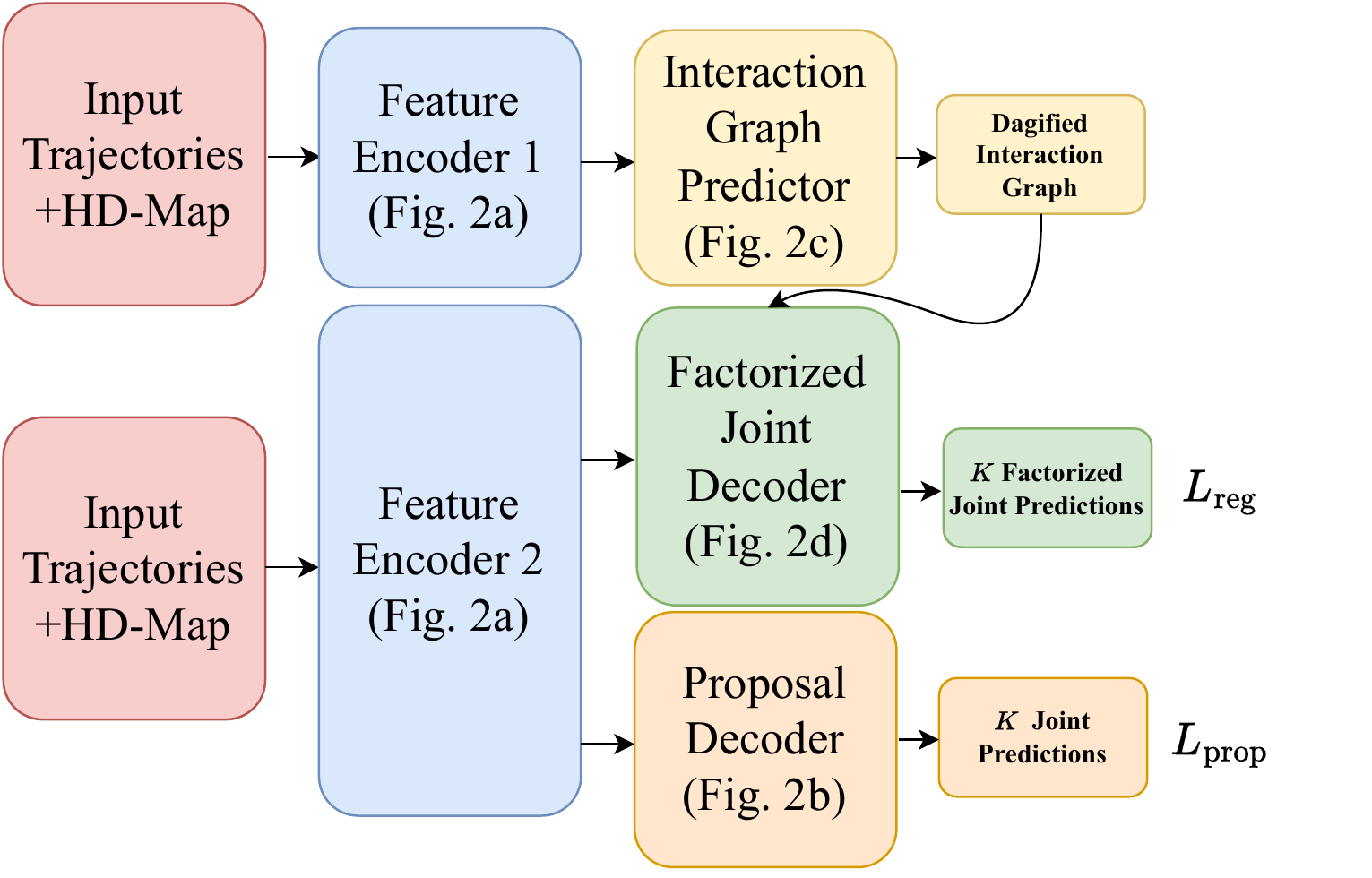}
    \caption{Factorized Joint Predictor training stage of FJMP.}
    \label{fig:trainingsystemdiagramstage2}
  \end{subfigure}
  \caption{High-level schematic of the training stages of FJMP.}
  \label{fig:fjmptrainingstages}
\end{figure*}

\Cref{fig:fjmptrainingstages} illustrates a high-level schematic of the FJMP architecture training stages at training time. We note that Feature Encoder 1 and Feature Encoder 2 consist of the same architecture as described in \cref{subsec:featureencoder}, but use separate weights. 

\vfill
\vfill

\subsection{Inference Time}
\label{app:inferencetime}

\begin{figure}[h]
    \centering
    \includegraphics[width=0.4\textwidth]{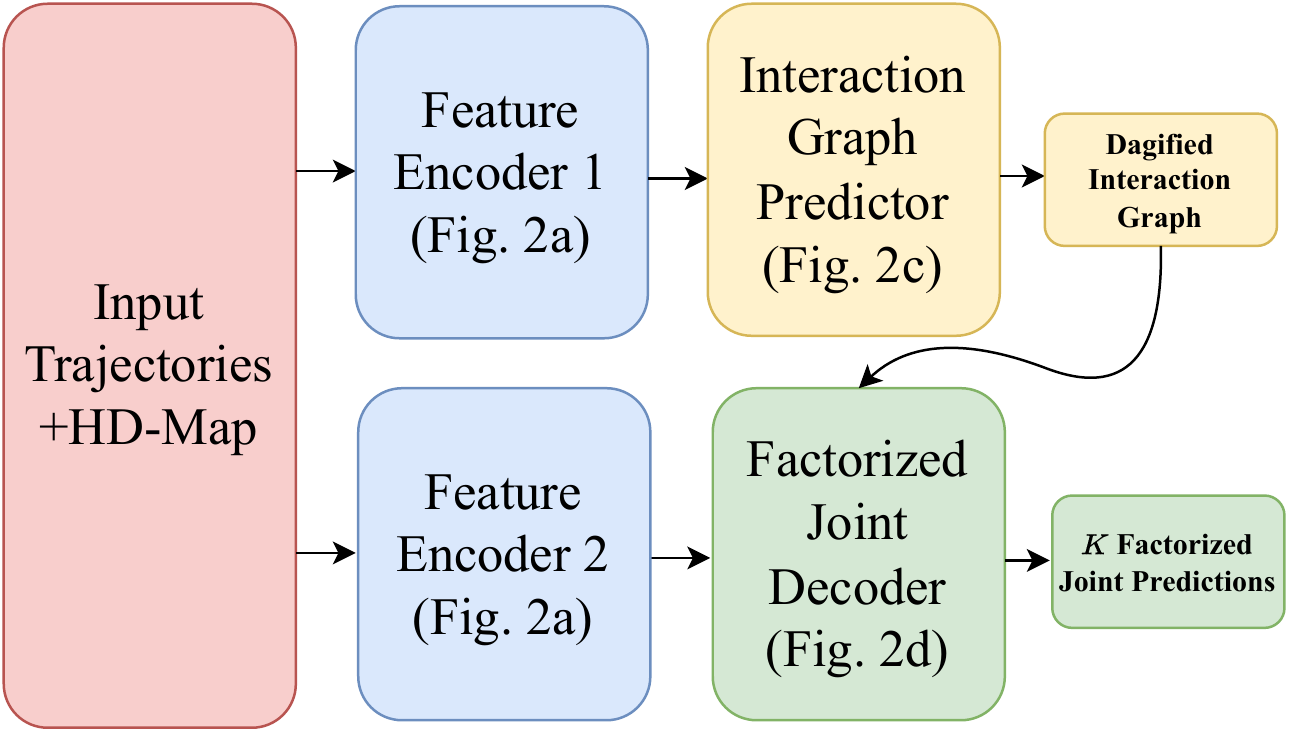}
    \caption{High-level schematic of the FJMP architecture at inference time.}
    \label{fig:inferencetime}
\end{figure}

\Cref{fig:inferencetime} illustrates a high-level schematic of the FJMP architecture and data flow at inference time. We note that at inference time the proposal decoders are removed.

\vfill
\vfill

\section{Non-Factorized Baseline}
\label{app:nonfactorizedbaseline}

We explain the non-factorized baseline described in \cref{sec:experiments} in more detail. The non-factorized baseline uses the same feature encoder architecture as FJMP, but the factorized joint decoder is replaced with a $\text{DECODE}$ module consisting of a residual block and linear layer for simultaneously decoding $K$ joint future trajectory coordinates, where diverse futures are obtained by appending a one-hot encoding of the modality index to the agent feature representation before feeding it into DECODE, as is done in FJMP. The $\text{DECODE}$ module is the same architecture as the $\text{DECODE}$ module used in FJMP. The non-factorized baseline is trained with the scene-level winner-takes-all smooth $\ell_1$ loss  $\mathcal{L}_{\text{reg}}$ that is described in \cref{app:lossfunctions}. The non-factorized baseline is trained with the same training hyperparameters as FJMP.

\section{Non-Factorized Baseline Ablation}
\label{app:nonfactorizedbaselineablations}

\begin{table}[h]
\centering
\resizebox{\columnwidth}{!}{
\begin{tabular}{l|ccc|cccc}
\toprule
Model & Multiple Futures Method & Hyperparameter Configuration & Feature Encoder & minFDE   & minADE & SMR  & SCR  \\ \midrule
LaneGCN \cite{liang2020lanegcn} & Separate Weights & LaneGCN & LaneGCN & 0.935 & 0.300 & 0.223 & 0.233 \\
- & One-hot Encoding & LaneGCN & LaneGCN & 0.807 & 0.264 & 0.142 & 0.010 \\ 
- & One-hot Encoding & FJMP & LaneGCN & 0.713 & 0.227 & 0.113 & 0.006 \\
Non-Factorized Baseline & One-hot Encoding & FJMP & FJMP & 0.643 & 0.199 & 0.088 & 0.004 \\ \bottomrule
\end{tabular}
}
\caption{Ablation study of the Non-Factorized Baseline model on the INTERACTION validation set. \textbf{Multiple Futures Method} denotes the method used to attain multiple joint futures. \textbf{Hyperparameter Configuration} denotes the hyperparameter settings for batch size, learning rate/step, and the number of training epochs. \textbf{Feature Encoder} denotes whether we use the LaneGCN feature encoder (LaneGCN) or the simplified LaneGCN feature encoder with fewer components (FJMP). }
\label{tab:nonfactorizedbaseline-ablation}
\end{table}

In \cref{tab:nonfactorizedbaseline-ablation}, we perform an ablation study on the various components of the non-factorized baseline model on the INTERACTION dataset. First, we ablate using a one-hot encoding for multiple futures (\textit{One-hot Encoding}) compared with using separate decoder weights for each joint future modality (\textit{Separate Weights}), as is done in LaneGCN \cite{liang2020lanegcn}. The one-hot encoding method significantly improves performance; this is because when using separate weights, the winner-takes-all training process quickly converges to one future joint modality, and thus the other decoders' weights never receive gradients for updating their weights. As a result, the collision rate (SCR) significantly improves when using the one-hot encoding method. Next, we ablate using the default hyperparameter configuration for LaneGCN compared with the FJMP hyperparameter configuration. Namely, LaneGCN trains for 36 epochs with a batch size of 128, with the learning rate decreasing by a factor of 10 at epoch 32. FJMP trains for 50 epochs with a batch size of 64, with the learning rate decreasing by a factor of 5 at epochs 40 and 48. The FJMP hyperparameter configuration significantly improves performance over the LaneGCN hyperparameter configuration. Finally, we ablate using the modified LaneGCN feature encoder (\textit{FJMP}) consisting of a GRU for processing agent trajectories instead of LaneGCN's proposed ActorNet module, 2 MapNet layers instead of 4, and the A2L and L2L blocks removed. These modifications yield further improvements in validation performance. 

\section{INTERACTION Ablation Study}
\label{app:interaction-ablation}

\begin{table}[h]
\centering
\resizebox{0.7\columnwidth}{!}{
\begin{tabular}{lcc|cc|cc}
\toprule
Model    & Prop? & TF? & minFDE   & minADE & iminFDE  & iminADE  \\ \midrule
Non-Factorized & \text{\ding{55}}  & \text{\ding{55}}  & 0.643 & 0.199 & 0.688 & 0.210 \\
FJMP     & \text{\ding{55}}  & \text{\ding{55}} & 0.647 & 0.200 & 0.690 & 0.212 \\
FJMP     & \text{\ding{55}}  & \text{\ding{51}}  & 0.644 & 0.200 & 0.688 & 0.212 \\ 
FJMP     & \text{\ding{51}} & \text{\ding{55}} & \underline{0.636} & \underline{0.197} & \underline{0.677} & \underline{0.208} \\
FJMP     & \text{\ding{51}} & \text{\ding{51}} &\textbf{0.630} & \textbf{0.194} & \textbf{0.671} & \textbf{0.206} \\ \bottomrule
\end{tabular}
}
\caption{Ablation study of FJMP on the INTERACTION validation set. \textbf{Prop?} denotes whether we include the proposal decoder during training. \textbf{TF?} denotes whether we teacher-force the influencer trajectories during training.}
\label{tab:interaction-ablation}
\end{table}

In \cref{tab:interaction-ablation}, we repeat the FJMP ablation study conducted in \cref{tab:argoverse2-ablation} on the INTERACTION dataset. The results are consistent with Argoverse 2, showing that both the proposal decoder and teacher forcing are critical for performance.

\section{Datasets}
\label{app:datasets}

\subsection{INTERACTION}
\label{app:interactiondataset}

INTERACTION requires predicting 3 seconds into the future given 1 second of past observations sampled at 10\,Hz. INTERACTION contains 47,584 training scenes, 11,794 validation scenes, and 2,644 test scenes. A scene consists of a 4\,s sequence of observations (1\,s past, 3\,s future) for each agent. INTERACTION contains pedestrians, bicyclists, and vehicles as context agents but only requires predicting vehicles in their multi-agent challenge.  As bounding box length/width information is not provided for the pedestrian/cyclist labels, we set the length and width to a pre-defined value of 0.7m. We note that pedestrians and cyclists are not differentiated in the INTERACTION dataset.

\subsection{Argoverse 2}
\label{app:argoverse2dataset}

Argoverse 2 requires predicting 6 seconds into the future given 5 seconds of past observations sampled at 10\,Hz. Argoverse 2 contains 199,908 training scenes and 24,988 validation scenes.  A scene consists of an 11\,s sequence of observations (5\,s past, 6\,s future) for each agent. Argoverse 2 requires predicting 5 agent types: vehicle, pedestrian, bicyclist, motorcyclist, and bus. As bounding box length/width information is not provided in the Argoverse 2 dataset, we use the following predefined length/width in meters for each agent type to construct the interaction labels (length/width): vehicle (4.0/2.0), pedestrian (0.7/0.7), bicyclist (2.0/0.7), motorcyclist (2.0/0.7), bus (12.5/2.5). 

\section{Training Details}
\label{app:trainingdetails}

\subsection{INTERACTION}
\label{app:interactiontrainingdetails}

The hidden dimension of FJMP is 128 except for the GRU history encoder, which has a hidden dimension of 256. The output of the GRU encoder is mapped to dimension 128 with a linear layer.  We set $K = 6$ for the factorized decoder and $K = 15$ for the proposal decoders. For training the interaction graph predictor, we set $\gamma = 5$ and $\alpha = [1, 2, 4]$. We set $\epsilon_I = 2.5$\,s. During training, we center and rotate the scene on a random agent, as an input normalization step. During validation and test time, we center and rotate the scene on the agent closest to the centroid of the agents' current positions. We use 2 MapNet layers, 2 L2A layers, and 2 A2A layers, where the L2A and A2A distance thresholds are set to 20\,m and 100\,m, respectively. We use all agents in the scene for context that contains a ground-truth position at the present timestep. As centerline information is not provided in INTERACTION, for each lanelet we interpolate $P$ evenly-spaced centerline points, where $P = \min\{10, \max \{L, R\}\}$ and $L, R$ are the number of points on the lanelet's left and right boundaries, respectively; that is, we restrict long lanelets to have a maximum of 10 evenly-spaced centerline points. At validation time, we consider for evaluation all vehicles that contain a ground-truth position at both the present and final timesteps. We train our model on the train and validation set with the same training hyperparameters before evaluating FJMP on the INTERACTION test set.

\subsection{Argoverse 2}
\label{app:argoverse2trainingdetails}

The details in \cref{app:interactiontrainingdetails} apply to Argoverse 2 with the following exceptions. For training the interaction graph predictor, we set $\gamma = 5$ and $\alpha = [1, 4, 4]$. We set $\epsilon_I = 6$\,s as interactions are comparatively more sparse in Argoverse 2. At validation time, we center on the ego vehicle. We increase the number of MapNet layers to 4 in Argoverse 2 to handle the larger amount of unique roadway. The L2A threshold is set to 10\,m as the centerline points are comparatively more dense in Argoverse 2 than in INTERACTION. We use all scored, unscored, and focal agents in the scene for context that contains a ground-truth position at the present timestep. In the \textit{Scored} validation setting (see \cref{tab:validation-results}), we consider for evaluation all scored and focal agents with a ground-truth position at both the present and final timesteps. In the \textit{All} validation setting (see \cref{tab:validation-results}), we consider for evaluation all scored, unscored, and focal agents with a ground-truth position at both the present and final timesteps. 

\section{Collision Checker}
\label{app:collisionchecker}

To construct the interaction labels as described in \cref{subsec:dagpredictor}, a collision checker is used to identify collisions between all pairs of timesteps in the future trajectories. We use the collision checker provided with the INTERACTION dataset. At each timestep, the collision checker defines each agent by a list of circles, and two agents are defined as colliding if the Euclidean distance between any two circles' origins of the given two agents is lower than the following threshold:
\begin{align}
    \epsilon_{C} := \frac{w_i + w_j}{\sqrt{3.8}},
\end{align}
where $w_i, w_j$ are the widths of agents $i, j$.

\section{Miss Rate}
\label{app:missrate}

\begin{table*}[t]
\centering
\resizebox{0.4\textwidth}{!}{
\begin{tabular}{cc|c}
\toprule
Actors Evaluated & Model  & $\text{SMR}_{\text{Argoverse2}}$ \\ \midrule
Scored & Non-Factorized     & 0.264\\
& FJMP & 0.259\\
& $\Delta$ & 0.005\\
All & Non-Factorized  & 0.259\\
& FJMP & 0.257\\
& $\Delta$ & 0.002\\
\bottomrule 
\end{tabular}
}
\caption{Non-Factorized Baseline vs. FJMP performance on Argoverse 2 SMR metric on the Argoverse 2 validation set. $\Delta$ denotes the difference in performance between FJMP and the Non-Factorized baseline.}
\label{tab:argoverse2miss}
\end{table*}

For both Argoverse 2 and INTERACTION, we use the definition of a miss used in the INTERACTION dataset: a prediction is considered a ``miss" if the longitudinal or latitudinal distance between the prediction and ground-truth endpoint is larger than their corresponding thresholds, where the latitudinal threshold is $\epsilon_{\text{lat}} := 1\,\text{m}$ and the longitudinal threshold is:
\begin{align}
    \epsilon_{\text{long}} := 
    \begin{cases}
        1, & \text{if $v \leq 1.4\,\text{m/s}$}\\
        1 + \frac{v-1.4}{11-1.4}, & \text{if $1.4\,\text{m/s} \leq v \leq 11\,\text{m/s}$}\\
        2, & \text{otherwise,}
    \end{cases}
\end{align}
where $v$ is the ground-truth velocity at the final timestep. We note that Argoverse 2 officially defines a miss as a prediction whose endpoint is more than 2\,m from the ground-truth endpoint; however, we report all miss rate numbers in \cref{tab:validation-results} using the miss rate definition in INTERACTION as it is a more robust measure of miss rate that takes into account the agent's velocity. For completeness, we report miss rate numbers for Argoverse 2 using the Argoverse 2 definition of a miss in \cref{tab:argoverse2miss}.

\section{Constant Velocity Model}
\label{app:constantvelocitymodel}

\begin{figure*}
  \centering
  \begin{subfigure}{0.8\textwidth}
    \includegraphics[width=1\textwidth]{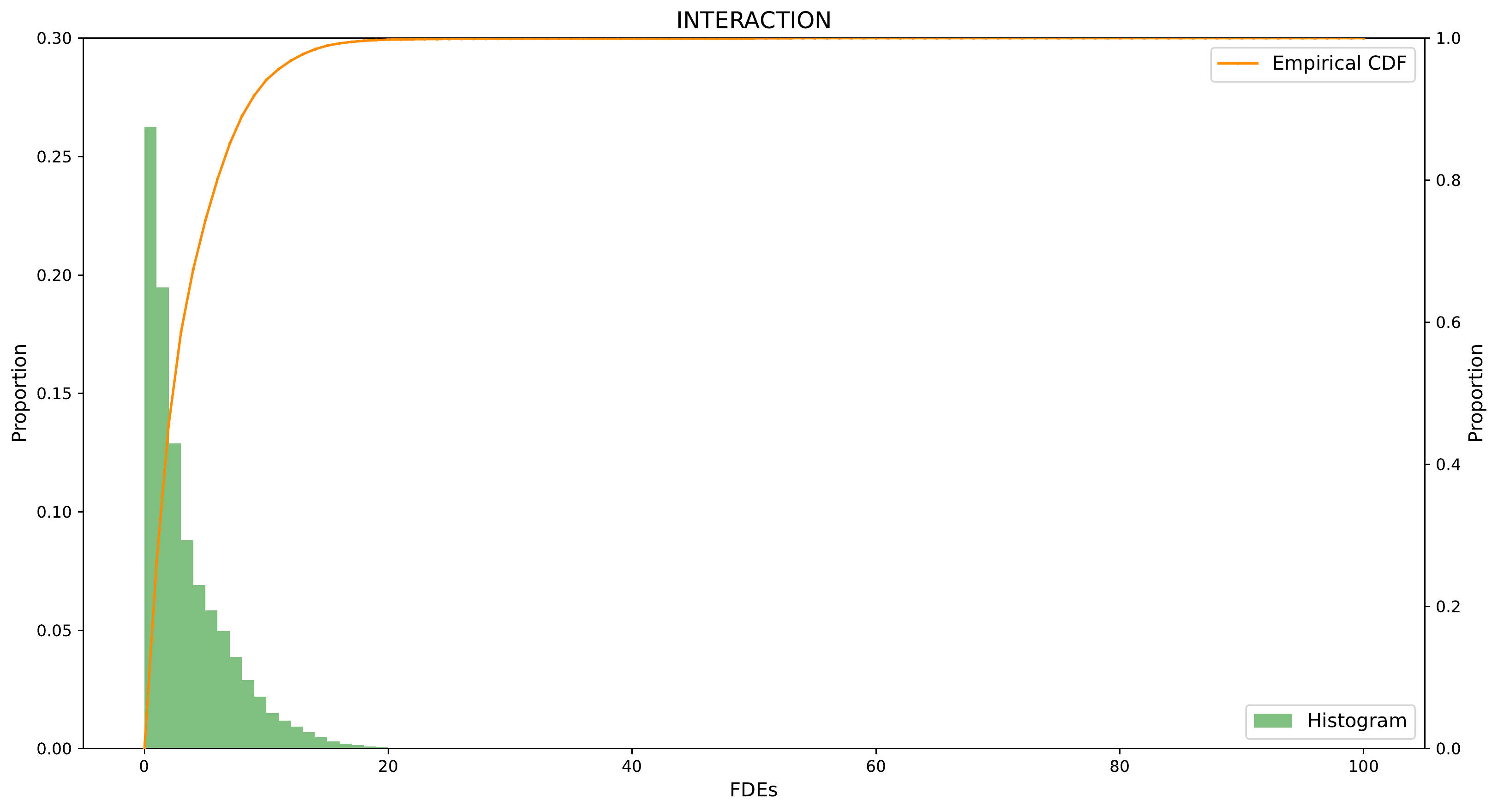}
    \caption{}
    \label{fig:histinteraction}
  \end{subfigure}
  \hfill
  \begin{subfigure}{0.8\textwidth}
    \includegraphics[width=1\textwidth]{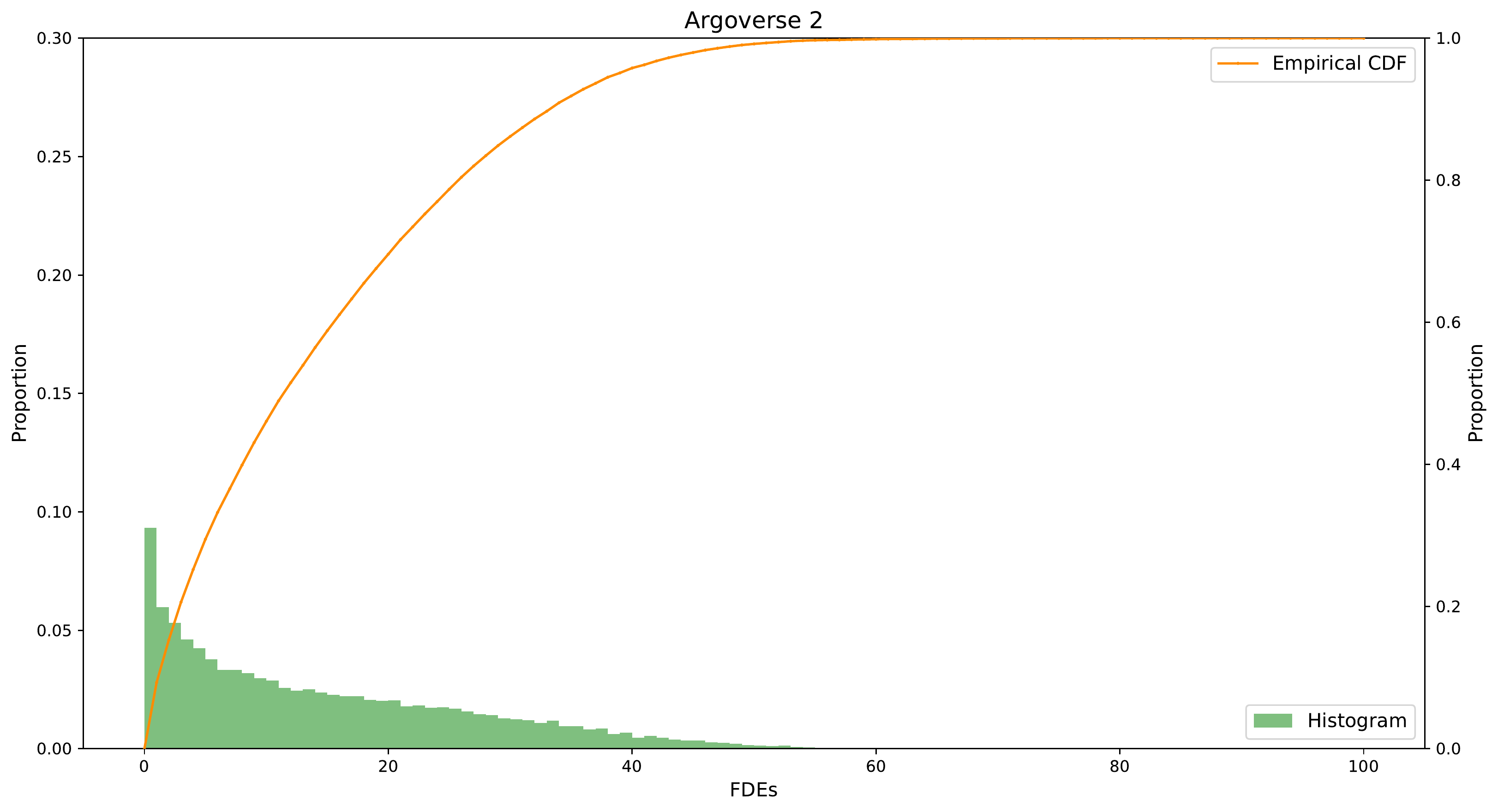}
    \caption{}
    \label{fig:histargoverse2}
  \end{subfigure}
  \caption{Histogram of FDEs on interacting agents in (a) the INTERACTION dataset, and (b) the Argoverse 2 dataset. The left y-axis corresponds to the histogram and the right y-axis corresponds to the empirical cumulative distribution function (CDF).}
  \label{fig:hist}
\end{figure*}

In \cref{sec:experiments}, we identify the kinematically complex interactive agents in the datasets by filtering for agents that attain at least $d$\,m in FDE with a constant velocity model. An interactive agent is defined as an agent with at least one incident edge in the ground-truth interaction graph, where $\epsilon_I = 2.5$\,s, as is explained in \cref{sec:experiments}. In this section, we describe the constant velocity model in more detail. The constant velocity model computes the average velocity over the observed timesteps and unrolls a future trajectory using the calculated constant velocity. Namely, the average velocity is calculated as:
\begin{align}
    \mathbf{v}_{\text{avg}} = \frac{1}{T_{\text{obs}}} \sum_{t \in [T_{\text{obs}}]} \mathbf{v}_t,
\end{align}
where $\mathbf{v}_t$ is the ground-truth velocity at timestep $t$. Using the constant velocity model, we calculate the agent-level FDE of all interactive agents in the INTERACTION and Argoverse 2 validation sets, respectively, where the FDE distributions are plotted in \cref{fig:hist}. We observe that a large proportion of the interactive agents have low FDE with a constant velocity model, especially in the INTERACTION dataset. By filtering out these kinematically simple agents, as is done in \cref{sec:experiments}, we can assess the model's joint prediction performance on agents that are both interactive and kinematically complex. In \cref{tab:constantvelocity}, we report the number of interactive agents in the INTERACTION and Argoverse 2 validation sets that attain at least $d$\,m in FDE, for $d=0,3,5$. We note that $d=0$ corresponds to the number of interactive agents in the respective validation sets.

\begin{table*}[t]
\centering
\resizebox{0.3\textwidth}{!}{
\begin{tabular}{cc|c}
\toprule
Dataset & $d$  & Count \\ \midrule
INTERACTION & 0     & 50967\\
(112994) & 3 & 21077\\
& 5 & 13069\\\midrule
Argoverse 2 & 0  & 37065 \\
(248719) & 3 & 29421 \\
& 5 & 26140\\
\bottomrule 
\end{tabular}
}
\caption{Number of \textit{interactive} agents in the INTERACTION and Argoverse 2 datasets that attain at least $d$\,m in FDE with a constant velocity model. In parentheses, we include the total number of evaluated agents (interactive + non-interactive) in the respective validation sets.}
\label{tab:constantvelocity}
\end{table*}

\section{FJMP vs. M2I Interaction Graphs}
\label{app:sparsevsdense}

\begin{figure*}
  \centering
  \begin{subfigure}{0.49\textwidth}
    \includegraphics[width=1\textwidth]{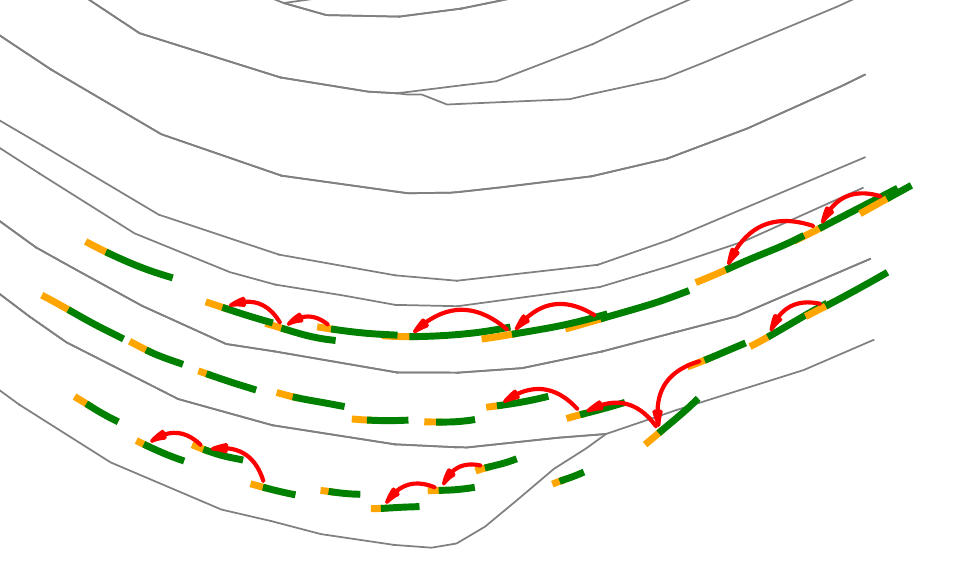}
    \caption{Interaction graph generated with FJMP labeling heuristic.}
    \label{fig:sparse}
  \end{subfigure}
  \hfill
  \begin{subfigure}{0.49\textwidth}
    \includegraphics[width=1\textwidth]{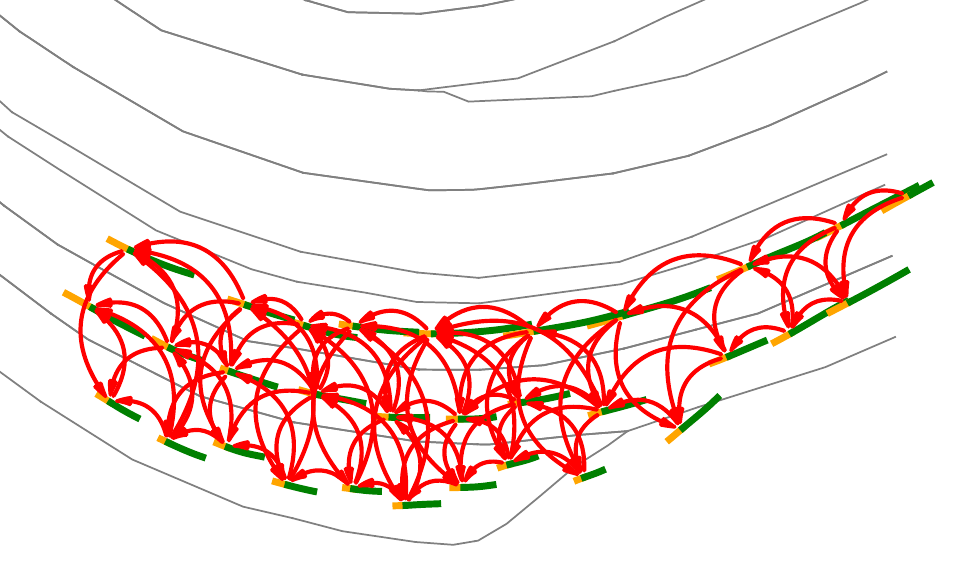}
    \caption{Interaction graph generated with M2I labeling heuristic.}
    \label{fig:dense}
  \end{subfigure}
  \caption{Comparison of FJMP and M2I labeling heuristics on a congested scene from the INTERACTION dataset. The ground-truth pasts are indicated in yellow and the ground-truth futures are indicated in green. Lane boundaries are depicted as grey lines. Each red arrow points from an influencer agent to its corresponding reactor agent. We note that two agents at the bottom-right of the scene are on the shoulder of the lane.}
  \label{fig:sparsevsdense}
\end{figure*}

\Cref{fig:sparsevsdense} illustrates the ground-truth interaction graph of a congested scene according to the FJMP and M2I heuristics, respectively. We observe that the M2I heuristic adds several superfluous edges, which would lead to unnecessary additional computation for the factorized decoder.

\section{Interaction Graph Predictor Performance}
\label{app:interactiongraphpredictorperformance}

\begin{table*}[t]
\centering
\resizebox{0.5\textwidth}{!}{
\begin{tabular}{cc|c}
\toprule
Dataset & Edge Type  & Edge Type Proportion \\ \midrule
INTERACTION & \texttt{no-interaction}  & 0.955 \\
& \texttt{m-influences-n}  & 0.037 \\
& \texttt{n-influences-m}  & 0.008\\\midrule
Argoverse 2 & \texttt{no-interaction}  & 0.973 \\
& \texttt{m-influences-n}  & 0.015 \\
& \texttt{n-influences-m}  & 0.013 \\
\bottomrule 
\end{tabular}
}
\caption{Edge type proportions in the INTERACTION and Argoverse 2 training set interaction graphs with the FJMP labeling heuristic.}
\label{tab:edgeproportion}
\end{table*}

\begin{table*}[t]
\centering
\resizebox{0.5\textwidth}{!}{
\begin{tabular}{cc|c}
\toprule
Dataset & Edge Type  & Edge Type Accuracy \\ \midrule
INTERACTION & \texttt{no-interaction}  & 0.992\\
& \texttt{m-influences-n}  & 0.940\\
& \texttt{n-influences-m}  & 0.939\\\midrule
Argoverse 2 & \texttt{no-interaction}  & 0.990\\
& \texttt{m-influences-n}  & 0.847\\
& \texttt{n-influences-m}  & 0.859\\
\bottomrule 
\end{tabular}
}
\caption{Accuracy of each edge type on the INTERACTION and Argoverse 2 validation sets with the FJMP interaction graph predictor.}
\label{tab:edgeaccuracies}
\end{table*}

\Cref{tab:edgeproportion} reports the proportion of \texttt{no-interaction}, \texttt{m-influences-n}, and \texttt{n-influences-m} edges in the INTERACTION and Argoverse 2 training sets. Due to the severe class imbalance, we employ a focal loss when training the interaction graph predictor, as explained in \cref{subsubsec:interactiongraphpredictor}. The edge type accuracies of the proposed interaction graph predictor on the INTERACTION and Argoverse 2 validation sets are reported in \cref{tab:edgeaccuracies}.

\subsection{Ground-truth Interaction Graph Performance}
\label{app:groundtruthinteractiongraph}

\Cref{tab:groundtruthig} compares the performance of FJMP with two modified versions of FJMP: (1) we replace the predicted interaction graphs at inference time with the ground-truth interaction graphs; and (2) we replace the predicted interaction graphs during training and inference time with the ground-truth interaction graphs. The results in \cref{tab:groundtruthig} indicate that the choice of interaction graph has a considerable effect on the performance of the factorized joint predictor, as indicated by an additional 4\,cm improvement in iminFDE with the ground-truth interaction graph at inference time over the predicted interaction graph. Moreover, when the model is trained and evaluated with the ground-truth interaction graphs, we see a substantial increase in performance over FJMP with the learned interaction graphs. This indicates that further refinement of the interaction graph predictor may yield additional performance improvements with our FJMP design, which we leave to future work.

\begin{table*}[t]
\centering
\resizebox{0.6\columnwidth}{!}{
\begin{tabular}{lcc|cc|cc}
\toprule
Model & Train IG & Inference IG         & minFDE   & minADE   & iminFDE  & iminADE  \\ \midrule
FJMP  & Learned & Learned & 1.963 & 0.812 & 3.204 & 1.273\\
FJMP  & Learned & Ground-truth & 1.947 & 0.807 & 3.165 & 1.265\\
FJMP  & Ground-truth & Ground-truth & 1.888 & 0.789 & 2.986 & 1.220 \\
%\multicolumn{2}{l|}{Improvement} & 0.069 & 0.019 &     & 0.167 & 0.042 \\ 
\bottomrule 
\end{tabular}
}
\caption{FJMP with ground-truth vs learned interaction graphs at training and inference time on the Argoverse 2 validation set, All setting. For each metric, the best model is \textbf{bolded}. \textbf{Train IG} indicates the interaction graphs that are used during training, where \textbf{Learned} denotes the predicted interaction graphs from the interaction graph predictor and \textbf{Ground-truth} denotes the interaction graphs obtained from the labeling heuristic. The \textbf{Inference IG} column is interpreted similarly.}
\label{tab:groundtruthig}
\end{table*}

\section{Qualitative Results}
\label{app:qualitativeresults}

\subsection{Argoverse 2}
\begin{figure}
    \centering
    \includegraphics[width=1.0\textwidth]{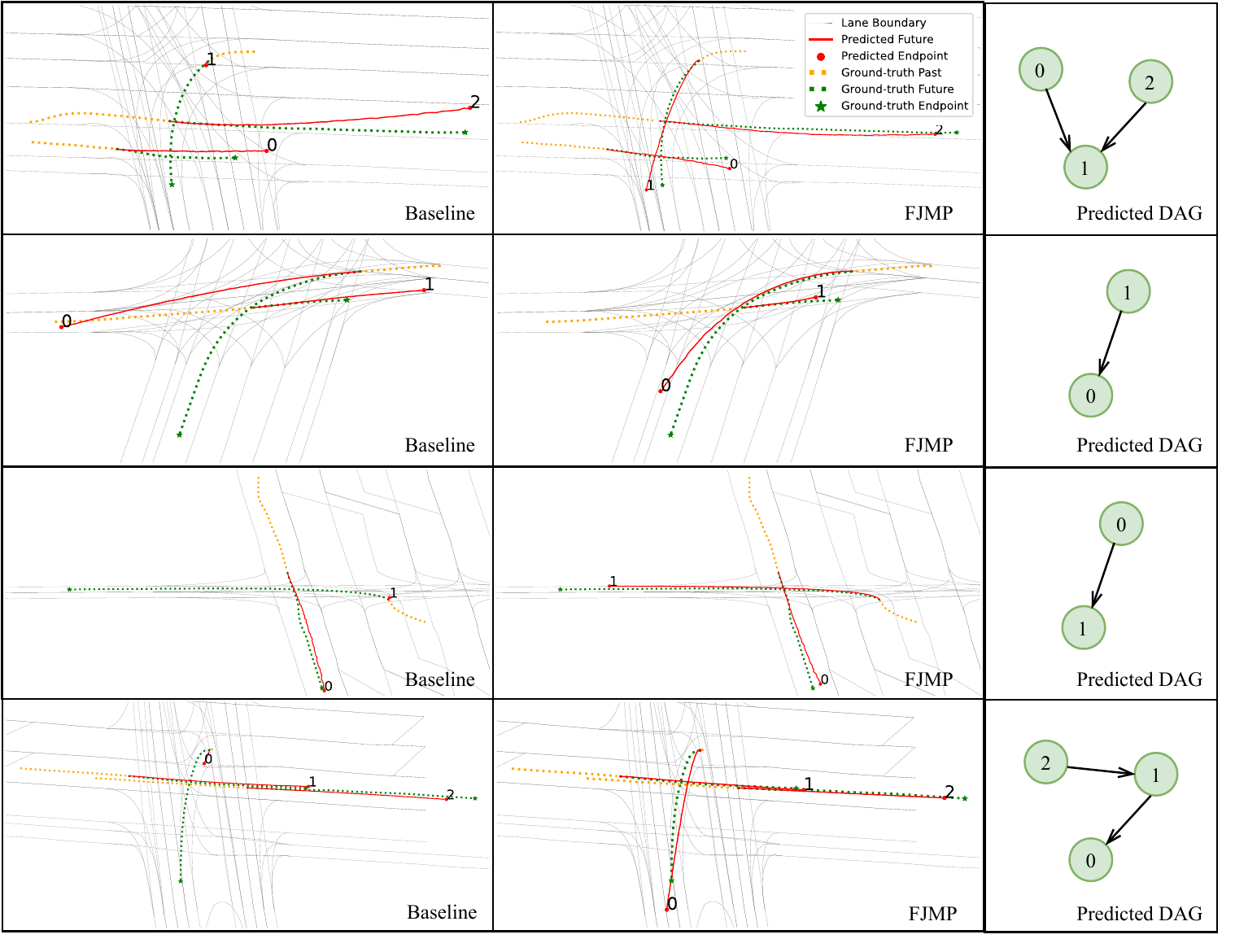}
    \caption{Qualitative examples of left-turn interactive scenes in the Argoverse 2 validation set. All predicted DAGs match the ground-truth DAG. In all scenes, FJMP correctly identifies the passing vehicle as the influencer and the left-turning vehicle as the reactor. The Non-Factorized baseline consistently predicts overly conservative behaviour that avoids the influencer trajectory. In contrast, FJMP consistently captures the proper left-turn behaviour. }
    \label{fig:passyield}
\end{figure}

\begin{figure}
    \centering
    \includegraphics[width=1.0\textwidth]{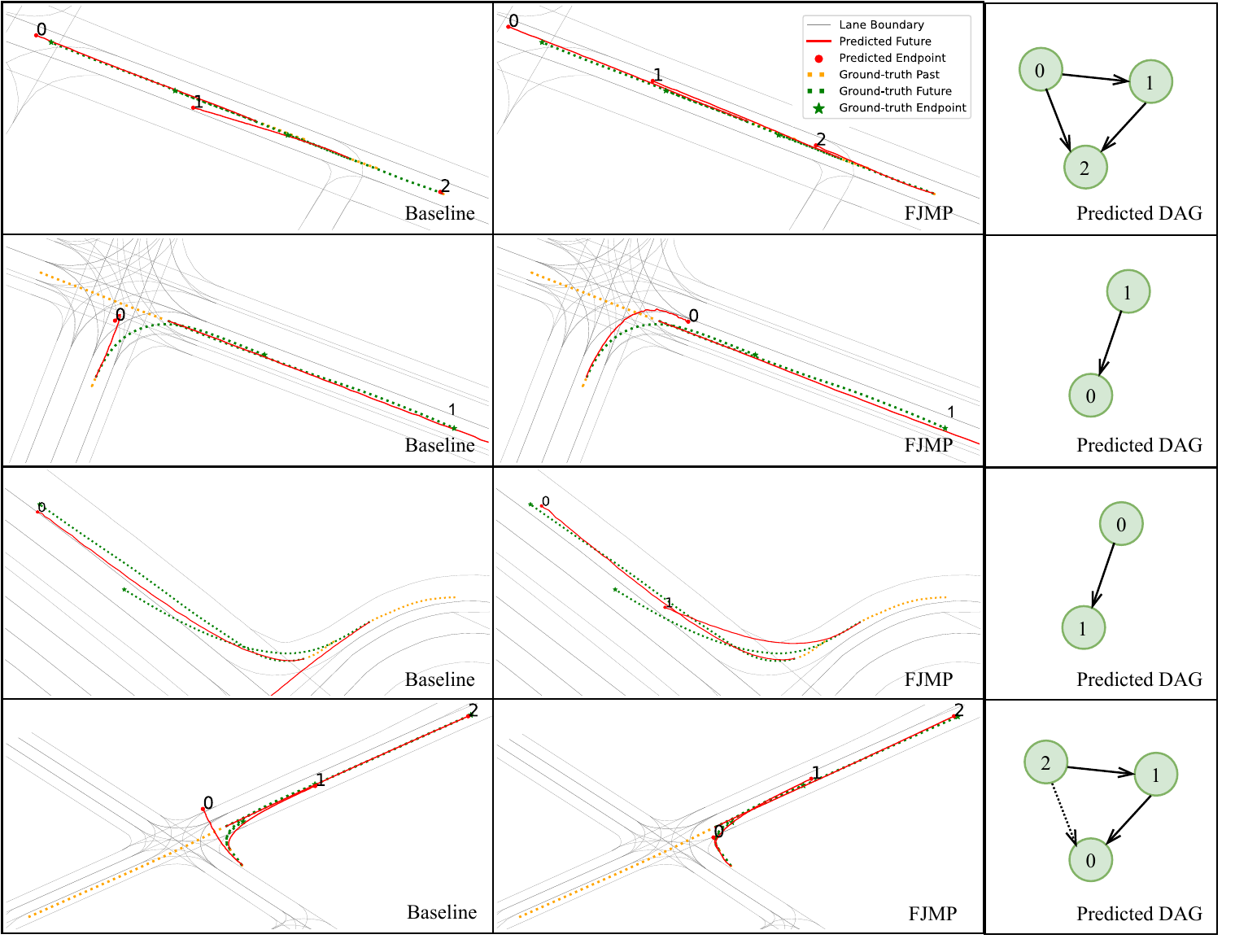}
    \caption{Qualitative examples of leader-follower interactive scenes in the Argoverse 2 validation set. Predicted DAGs are shown on the right, where true positive edges are indicated in solid black and true negative edges are shown in dotted black. In all of the above scenes, FJMP correctly predicts chains of influencer-reactor relationships. In the first row, the non-factorized baseline predicts conservative behaviour for the trailing vehicle. In contrast, FJMP predicts proper leader-follower behaviour for the trailing vehicle (leaf node in the DAG). In the second and third rows, the right-turn mode of the trailing vehicle is missed by the non-factorized baseline, whereas FJMP correctly identifies the right-turn mode due to correctly identifying the leader-follower interaction. In the last row, the non-factorized baseline predicts scene-incompliant behaviour for the trailing vehicle whereas FJMP predicts proper leader-follower dynamics reflecting the predicted DAG.}
    \label{fig:leaderfollower}
\end{figure}

\begin{figure}
    \centering
    \includegraphics[width=0.6\textwidth]{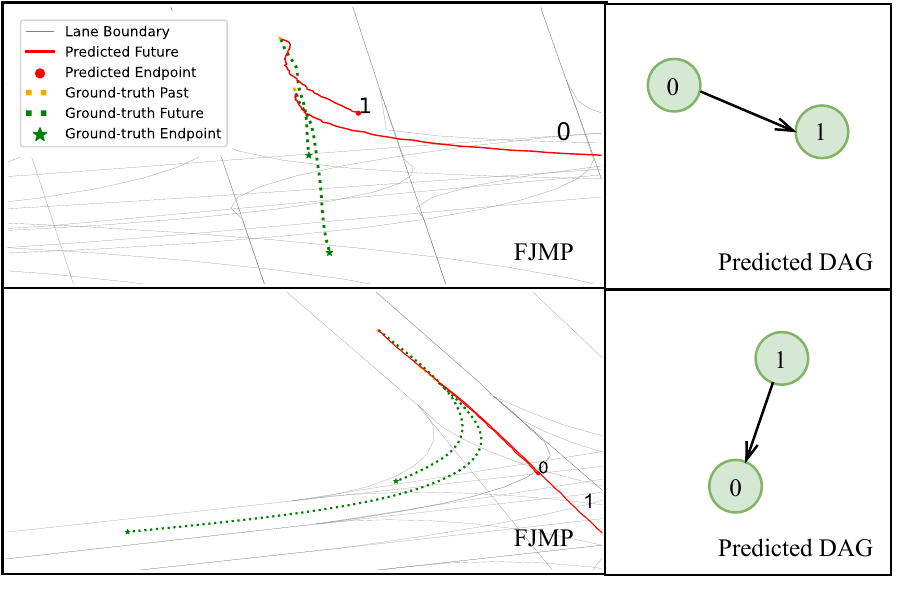}
    \caption{Qualitative examples of failure cases of the FJMP model. All predicted DAGs match the ground truth. In both rows, the interaction graph is correctly predicted; however, the influencer trajectory is erroneously predicted, which negatively biases the reactor's prediction to follow the influencer.}
    \label{fig:failurecases}
\end{figure}

In this section, we show qualitative results on scenes in the Argoverse 2 validation set where we show side-by-side comparisons between FJMP and the Non-Factorized Baseline. In \cref{fig:passyield} and \cref{fig:leaderfollower}, for each row, the left panel shows the non-factorized baseline predictions, the middle panel shows FJMP predictions, and the right panel shows the predicted DAG. We visualize only the best scene-level modality to avoid clutter.  In \cref{fig:passyield}, we show examples where FJMP reasons properly in scenes with interactive pass-yield behaviours. In contrast, the non-factorized baseline incorrectly predicts conservative behaviour where the yielding vehicle avoids the passing vehicle's trajectory. In \cref{fig:leaderfollower}, we show qualitative examples where FJMP correctly identifies chains of leader-follower interactions, which in turn leads to more accurate leader-follower predictions than the non-factorized baseline. In \cref{fig:failurecases}, we illustrate two failure cases of the FJMP model. In both cases, an erroneous influencer future prediction negatively biases the downstream reactor prediction.

\subsection{INTERACTION}

\begin{figure}
    \centering
    \includegraphics[width=0.6\textwidth]{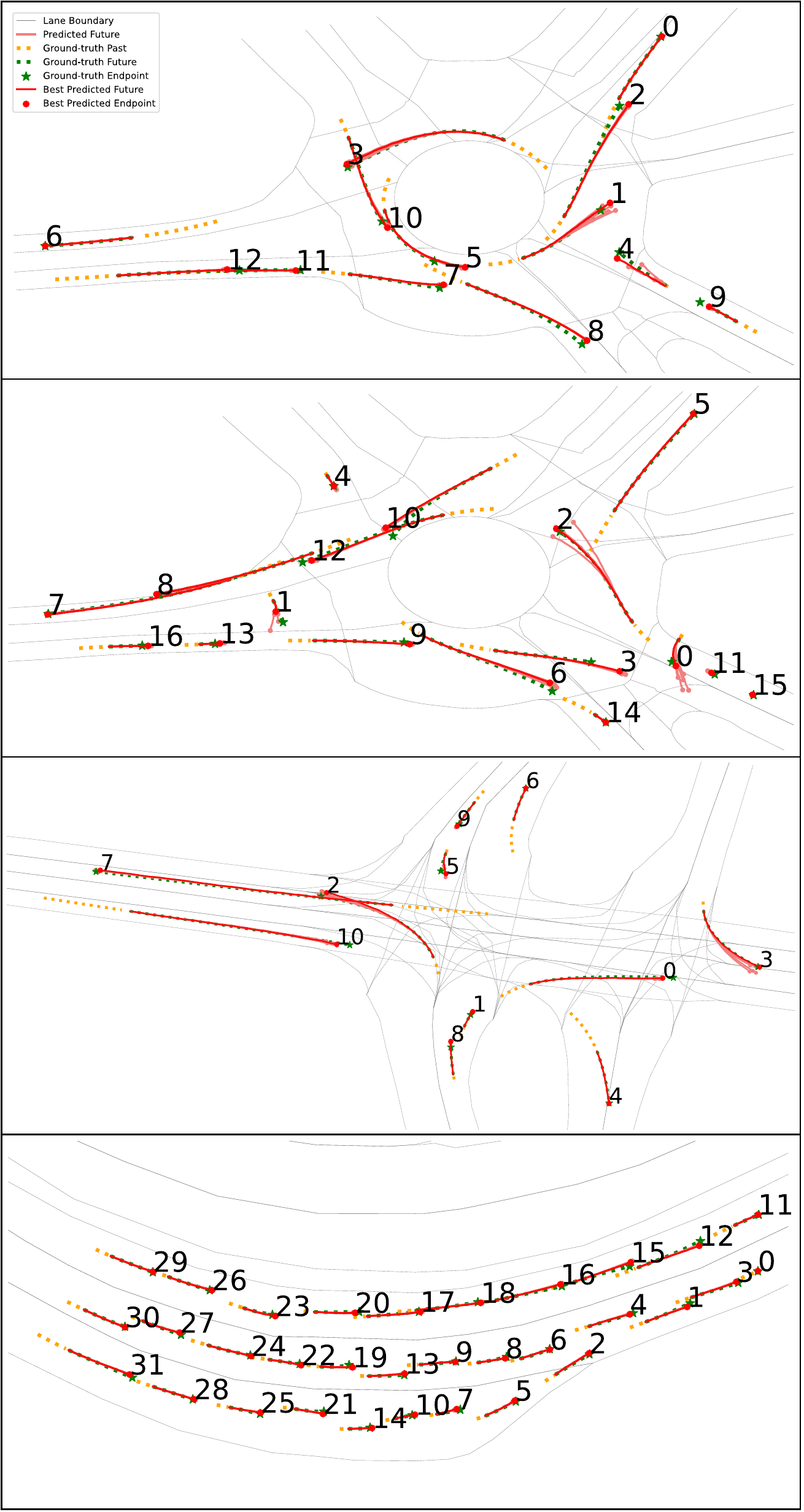}
    \caption{Qualitative examples of FJMP on agent-dense scenes in the INTERACTION dataset.}
    \label{fig:interactionvisuals}
\end{figure}

\Cref{fig:interactionvisuals} shows qualitative results of FJMP on various scenes in the INTERACTION dataset, with all $K = 6$ scene-level modalities visualized. We emphasize FJMP's ability to produce accurate and scene-consistent predictions for scenes with a large number of interacting agents.

\clearpage

\end{document}